\documentclass[10pt,journal,compsoc]{IEEEtran}
\usepackage{times}
\usepackage{epsfig}
\usepackage{graphicx}
\usepackage{amsmath}
\usepackage{amssymb}
\usepackage{xcolor}
\usepackage{tabu}
\usepackage{booktabs}
\usepackage{multirow}
\usepackage{algorithm}
\usepackage{algorithmic}
\usepackage[utf8]{inputenc}
\usepackage{makecell}

\providecommand{\tabularnewline}{\\}
\usepackage{xspace}

\makeatletter
\DeclareRobustCommand\onedot{\futurelet\@let@token\@onedot}
\def\@onedot{\ifx\@let@token.\else.\null\fi\xspace}

\def\eg{\emph{e.g}\onedot}

\def\wrt{w.r.t\onedot} 
\def\etal{\emph{et al}\onedot}
\makeatother

\ifCLASSOPTIONcompsoc
  \usepackage[nocompress]{cite}
\else
  \usepackage{cite}
\fi
\ifCLASSINFOpdf
\else
\fi
\usepackage{hyperref}
\usepackage{url}

\begin{document}
%
\title{Pixel2Mesh++: 3D Mesh Generation and Refinement from Multi-View Images}
%
%
%
%

\author{Chao Wen$^\star$,
        Yinda Zhang$^\star$,
        Chenjie Cao,
        Zhuwen Li,
        Xiangyang Xue,
        Yanwei Fu$^\dag$
        
\thanks{$^\star$ indicates equal contributions, $^\dag$ refers to the corresponding author.}

\IEEEcompsocitemizethanks{\IEEEcompsocthanksitem Wenchao,  Chenjie Cao, Xiangyang Xue, and Yanwei Fu are with Fudan University, China.  E-mail: woonchao@163.com, \{xyxue,  yanweifu\}@fudan.edu.cn.
\IEEEcompsocthanksitem Yinda Zhang is with Google. Email: zhangyinda@gmail.com.
\IEEEcompsocthanksitem Zhuwen Li is with Nuro, Inc., Mountain View, California. 
Email: lzhuwen@gmail.com.
\IEEEcompsocthanksitem Yanwei Fu is with School of Data Science, Fudan University.  Email: yanweifu@fudan.edu.cn.
}
}

\IEEEtitleabstractindextext{%
\begin{abstract}
We study the problem of shape generation in 3D mesh representation from a small number of color images with or without camera poses. 
While many previous works learn to hallucinate the shape directly from priors, we adopt to further improve the shape quality by leveraging cross-view information with a graph convolution network.
Instead of building a direct mapping function from images to 3D shape, our model learns to predict series of deformations to improve a coarse shape iteratively.
Inspired by traditional multiple view geometry methods, our network samples nearby area around the initial mesh's vertex locations and reasons an optimal deformation using perceptual feature statistics built from multiple input images.
Extensive experiments show that our model produces accurate 3D shapes that are not only visually plausible from the input perspectives, but also well aligned to arbitrary viewpoints.
With the help of physically driven architecture, our model also exhibits generalization capability across different semantic categories, and the number of input images. Model analysis experiments show that our model is robust to the quality of the initial mesh and the error of camera pose, and can be combined with a differentiable renderer for test-time optimization.
\end{abstract}

}

\maketitle

\IEEEdisplaynontitleabstractindextext

%
\IEEEpeerreviewmaketitle

\IEEEraisesectionheading{\section{Introduction}\label{sec:introduction}}
\IEEEPARstart{T}{he} topic of shape generation has become increasingly popular over the past decade. 
With the astonishing capability of deep learning, lots of works have been demonstrated to successfully generate the 3D shape from merely a single color image.
However, due to limited visual cues from only one viewpoint, single image based approaches usually produce rough geometry in the occluded area and do not perform well when generalized to domains other than training, \eg cross semantic categories.

\begin{figure}[t]
	\centering
    	\includegraphics[width=\columnwidth]{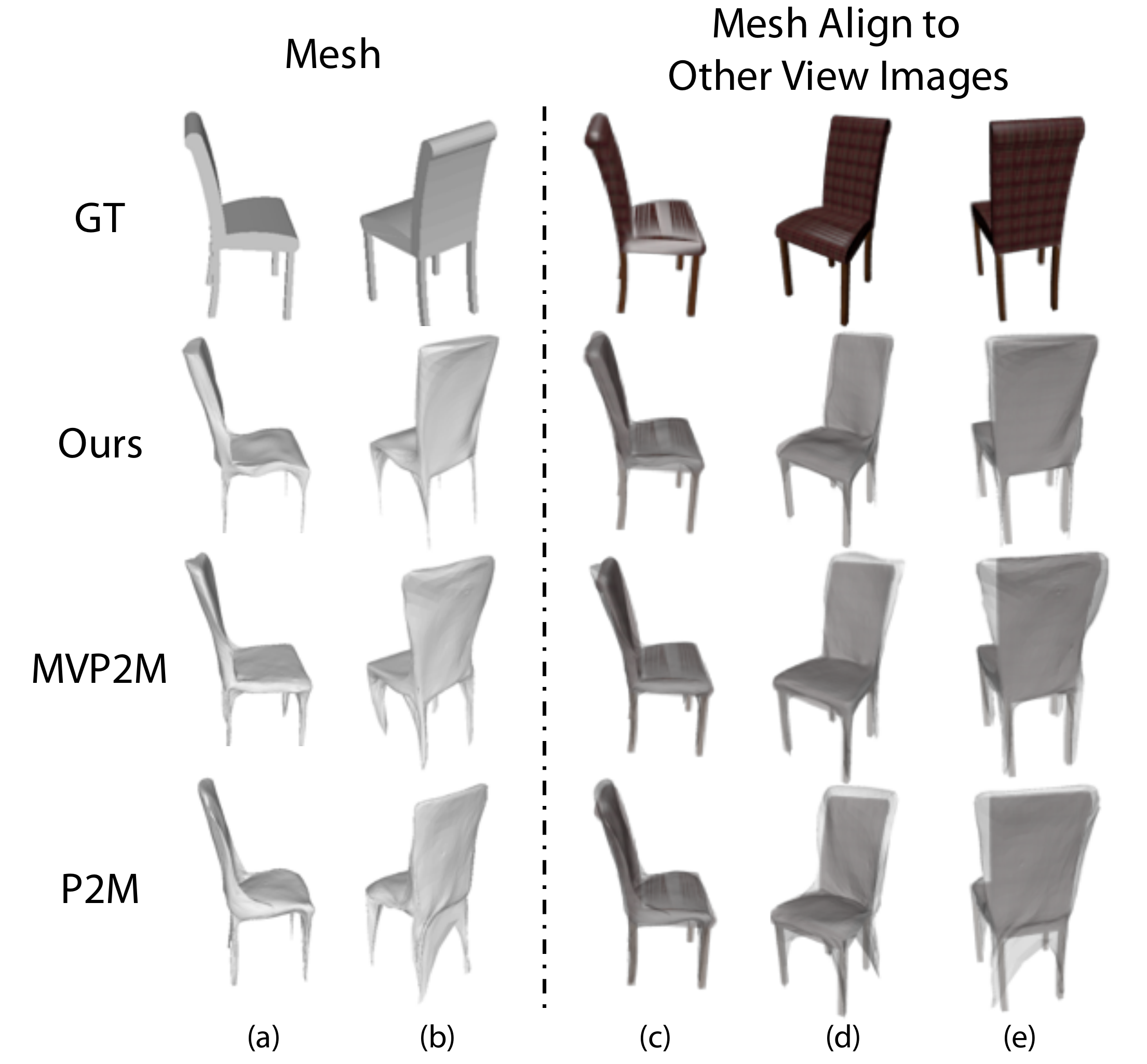}
	\caption{{\bf Multi-View Shape Generation.} From multiple input images, we produce shapes aligning well to input (c and d) and arbitrary random (e) camera viewpoint.
	Single view based approach, \eg Pixel2Mesh (P2M) \cite{wang2018pixel2mesh}, usually generates shape looking good from the input viewpoint (c) but significantly worse from others.
	Naive extension with multiple views (MVP2M, Sec. \ref{sec:mvp2m}) does not effectively improve the quality.}
	\label{fig:teaser}
\end{figure}

Adding a few more multi-view images (\eg 3-5) of the object is an effective way to provide the shape generation system with more information about the 3D shape.
On one hand, multi-view images provide more visual appearance information, and thus grant the system with more chance to build the connection between 3D shape and image priors.
On the other hand, it is well-known that traditional multi-view geometry methods \cite{HarlteyZ2001} accurately infer 3D shape from correspondences across views, which is analytically well defined and less vulnerable to the generalization problem. 
Even though typical multiview methods suffer from many problems, such as large baselines and textureless regions, and are likely to degenerate with very limited input images (\eg less than 5), the way they leverage cross-view information might be implicitly encoded and learned by a deep model.
While well-motivated, there are very few works in the literature exploiting in this direction. And a na\"ive multi-view extension that fuses output 3D shapes or intermediate image features of single image based model does not work well as shown in Fig. \ref{fig:teaser}.

In this work, we propose a deep learning model to generate the object shape from multiple color images with or without known camera poses.
Especially, we focus on endowing the deep model with the capacity of improving shapes using cross-view information.
The key component in our method is a new network architecture, named Multi-View Deformation Network (MDN), which works in conjunction with the Graph Convolutional Network (GCN) architecture proposed in Pixel2Mesh \cite{wang2018pixel2mesh} to generate accurate 3D geometry shape in the desirable mesh representation.
In Pixel2Mesh, a GCN is trained to deform an initial shape to the target using features from a single image, which often produces plausible shapes but lack of accuracy (Fig. \ref{fig:teaser} P2M).
We inherit this characteristic of ``generation via deformation'' and further deform the mesh in MDN using features carefully pooled from multiple images.
Instead of learning to hallucinate via shape priors like in Pixel2Mesh, MDN reasons shapes according to correlations across different views through a physically driven architecture inspired by classic multi-view geometry methods.
In particular, MDN proposes hypothesis deformations for each vertex and move it to the optimal location that best explains features pooled from multiple views.
By imitating correspondences search rather than learning priors, MDN generalizes well in various aspects, such as cross semantic category, number of input views, and the mesh initialization.


Besides the above-mentioned advantages, MDN is in addition featured with several good properties.
First, it can be trained end-to-end. Note that it is non-trivial since MDN searches deformation from hypotheses, which requires a non-differentiable argmax/min.
Inspired by \cite{KendallMDH17}, we apply a differentiable 3D soft argmax, which takes a weighted sum of the sampled hypotheses as the vertex deformation. Consequently, this component becomes differentiable and the whole system can be trained end-to-end.
Second, it works with varying number of input views in a single forward pass. This requires the feature dimension to be invariant with the number of inputs, which is typically broken when aggregating features from multiple images (\eg when using concatenation).

To address this issue, we concatenate the statistics (\eg mean, max, and standard deviation) of the pooled feature, which achieves invariance against the order of the input images as well.
We find this statistics feature encoding explicitly provides the network cross-view information, and encourages it to automatically utilize image evidence when more are available.
In addition, it can cope with the varying initial mesh topology. Thanks to the strategy of separating the generation of 3D shapes based on image priors and the shape refinement, MDN can refine the initial mesh created from various other sources, \eg Pixel2Mesh~\cite{wang2018pixel2mesh}, or via implicit representation method such as the Deep Implicit Surface Network (DISN)~\cite{NIPS2019_DISN}.
We found that MDN can still further improve the accuracy of the 3D shape even when the initial mesh has stronger expressive ability.
Moreover, we found that our method is robust to noisy camera pose, and can still work predicted camera pose if it is not known beforehand.
In this case, we estimate camera parameters which bring an object from the world coordinates to the image coordinates through a camera pose estimation network. This relaxation on the input made MDN more convenient to use in real scenario where camera pose may not be immediately accessible.
Last but not least, the nature of ``generation via deformation'' allows an iterative refinement. 
In particular, the model output can be taken as the input, and the quality of the 3D shape is gradually improved throughout iterations.
With these desiring features, our model achieves the state-of-the-art performance on ShapeNet for shape generation from multiple images under standard evaluation metrics.

To summarize, we propose a GCN framework that produces 3D shape in mesh representation from a small number of observations of the object in different viewpoints.
The core component is a physically driven architecture that searches optimal deformation to improve a coarse mesh using perceptual feature statistics built from multiple images, which produces accurate 3D shape and generalizes well across different semantic categories, numbers of input images, and the quality of coarse meshes. Our method can produce 3D shapes even when the initial topology is varied and with or without knowing the ground truth camera position.

An early version of this work has been published in~\cite{wen2019pixel2mesh++}. Compared with~\cite{wen2019pixel2mesh++}, we extend our method to work without known camera pose, with various initialization methods, more experiments and model analysis, as well as other technical improvements such as differentiable renderer.
We show that our method can further deform  the initial coarse meshes from those objects of arbitrary topology.

\section{Related Work}

\subsection{3D Shape Representations}
Since 3D CNN is readily applicable to 3D volumes, the volume representation has been well-exploited for 3D shape analysis and generation \cite{choy20163d,wang2017cnn}. With the debut of PointNet \cite{qi2016pointnet}, the point cloud representation has been adopted in many works \cite{fan2017point,QiLWSG18}. Most recently, the mesh representation \cite{KatoUH18,wang2018pixel2mesh} has become competitive due to its compactness and nice surface properties. Some other representations have been proposed, such as geometry images \cite{SinhaUHR17}, depth images \cite{TatarchenkoDB16,Richter018}, classification boundaries \cite{mescheder2019occupancy,chen2019learning,Liao2018CVPR, atzmon2019controlling,SAL_Atzmon_2020_CVPR}, signed distance function \cite{Park_2019_CVPR}, etc., and most of them require post-processing to get the final 3D shape. Consequently, the shape accuracy may vary and the inference take extra time.

\subsection{Single view shape generation}
Classic single view shape reasoning can be traced back to shape from shading \cite{DurouFS08,ZhangTCS99}, texture \cite{MarinosB90}, and de-focus \cite{FavaroS05}, which only reason the visible parts of objects. 
With deep learning, many works leverage the data prior to hallucinate the invisible parts, and directly produce shape in 3D volume \cite{choy20163d,GirdharFRG16,WuZXFT16,HaneTM17,RieglerUG17,TatarchenkoDB17,JohnstonGCR17}, point cloud \cite{fan2017point}, mesh models \cite{KatoUH18,groueix2018AtlasNet,goel2020shape, ye2021shelf,gupta2020neuralmeshflow}, or as an assembling of shape primitive \cite{TulsianiSGEM17,Niu0018,paschalidou2019superquadrics,genova2019learning,deng2020cvxnet,paschalidou2021neural}.
Alternatively, 3D shape can be also generated by deforming an initialization, which is more related to our work.
Tulsiani \etal\cite{TulsianiKCM17} and Kanazawa \etal\cite{KanazawaTEM18} learn a category-specific 3D deformable model and reasons the shape deformations in different images. 
Wang \etal \cite{wang2018pixel2mesh} learn to deform an initial ellipsoid to the desired shape in a coarse to fine fashion.
Combining deformation and assembly, Huang \etal\cite{HuangWK15} and Su \etal\cite{SuHMKG14} retrieve shape components from a large dataset and deform the assembled shape to fit the observed image. 
Kuryenkov \etal \cite{kurenkov2018deformnet} learns free-form deformations to refine shape.
Even though with impressive success, most single view deep models adopt an encoder-decoder framework, and it is arguable if they perform shape generation or shape retrieval \cite{TatarchenkoRRLKB19}.
\textcolor{black}{Groueix~\etal~\cite{groueix2018AtlasNet} also learn to generate 3D shapes by deforming explicit surfaces. Their method dubbed Atlasnet, uses the global latent vector and MLP parameters to represent the patches deformation.
In contrast, our work models the deformation by passing message in GCN and gathers local perceptual feature for each vertex of GCN.
It is feasible for Atlasnet to model the shape with holes by using multiple patches, while it is non-trivial to handle patch overlap and get the continuous surface meshes.
Our architecture, however, allows our network to generate and refine plausible smooth shapes when proper coarse shape initialization is used.}

\subsection{Multi-view shape generation}
Recovering 3D geometry from multiple views has been well studied. 
Traditional multi-view stereo (MVS) \cite{HarlteyZ2001} relies on correspondences built via photo-consistency and thus it is vulnerable to large baselines, occlusions, and texture-less regions. 
Most recently, deep learning based MVS models have drawn attention, and most of these approaches \cite{yao2018mvsnet, huang2018deepmvs, im2018dpsnet, zhang2018activestereonet} rely on a cost volume built from depth hypotheses or plane sweeps. 
Donne \etal~\cite{donne2019learning} refine a set of input depth maps in the image domain.
However, these approaches usually generate depth maps, and it is non-trivial to fuse a full 3D shape from them. 
On the other hand, direct multi-view shape generation uses fewer input views with large baselines, which is more challenging and has been less addressed. 
Choy \etal\cite{choy20163d} propose a unified framework for single and multi-view object generation reading images sequentially. Kar \etal\cite{kar2017learning} learn a multi-view stereo machine via recurrent feature fusion. Gwak \etal\cite{GwakCCGS17} learns shapes from multi-view silhouettes by ray-tracing pooling and further constrains the ill-posed problem using GAN. 
Our approach belongs to this category but is fundamentally different from the existing methods. Rather than sequentially feeding in images, our method learns a GCN to deform the mesh using features pooled from all input images at once.

\textcolor{black}{
\subsection{Differentiable Rendering}
Differentiable rendering retains derivatives during the forward process of image synthesis~\cite{loper2014opendr}. Therefore, many differentiable rendering enhanced neural networks can directly optimize their neural fields~\cite{xie2021neural} through image pixel colors, i.e., neural rendering.
}
\textcolor{black}{
According to the representation, this type of work can be categorized into neural radiance fields~\cite{mildenhall2020nerf}, implicit differentiable renderer~\cite{niemeyer2020differentiable,yariv2020multiview} and differentiable rasterizer~\cite{liu2020general}. Neural radiance fields (NeRF) and its successors~\cite{mildenhall2020nerf,liu2020neural,barron2021mipnerf,yu2021pixelnerf,song2019autoint,grf2020,henzler2021unsupervised} focus on representing the scene as 5D neural radiance fields. The success of these approaches mostly requires per-scene optimization, which is more suitable for novel-view synthesis tasks than 3D shape generation.
Among these, Yu~\etal~\cite{yu2021pixelnerf} propose pixelNeRF that shows the potential of cross-category and sparse input view rendering. However it is difficult to extract a high-quality surface. Implicit differentiable renderer approaches~\cite{niemeyer2020differentiable,yariv2020multiview} can generate 3D shapes via 2D supervision and volumetric rendering. Liu~\etal~\cite{liu2020general} propose a soft version rasterizer and learn to deform template mesh using multiple silhouette and color images.
Despite that these approaches could produce compelling results having dense input views, they still have performance degradation when very few views are available.
Our method can not only handle this this situation, but also can absorb differentiable renderer as a tool to optimize 3D shapes during inference.
}

\subsection{Camera Pose Estimation}
As a sub-problem in 3D shape recovery, many works are trying to tackle the problem of camera pose estimation.
Kendall \etal~\cite{kendall2015posenet} proposed PoseNet, which represents the pose as the camera position and camera rotation, and obtains the value of the pose by normalizing the output quaternion. 
Since two quaternions with opposite signs represent the same rotation under boundary condition, Wu \etal~\cite{wu2017delving} use Euler angles to represent rotation. 
Peng \etal~\cite{peng2019pvnet} regard camera pose estimation as a key point regression problem, and use voting strategy to densely estimate the key point offset. 
Recently, Zhou \etal~\cite{zhou2019continuity} show that the discontinuity of the rotation representation is the main reason for the difficulty in pose estimation, and proposed a continuous 6D rotation representation. 
Xu \etal~\cite{NIPS2019_DISN} continue to use the 6D rotation representation, and convert the pose numerical regression to the matching problem of the point cloud, which enhances the generalization performance.
We followed the 6D rotation representation, and further represented the translation as the scalar offset of the object along the optical axis in the camera coordinate.


\section{Method}

\subsection{System Overview}

\begin{figure*}[h!]
	\centering
	\includegraphics[width=1.8\columnwidth]{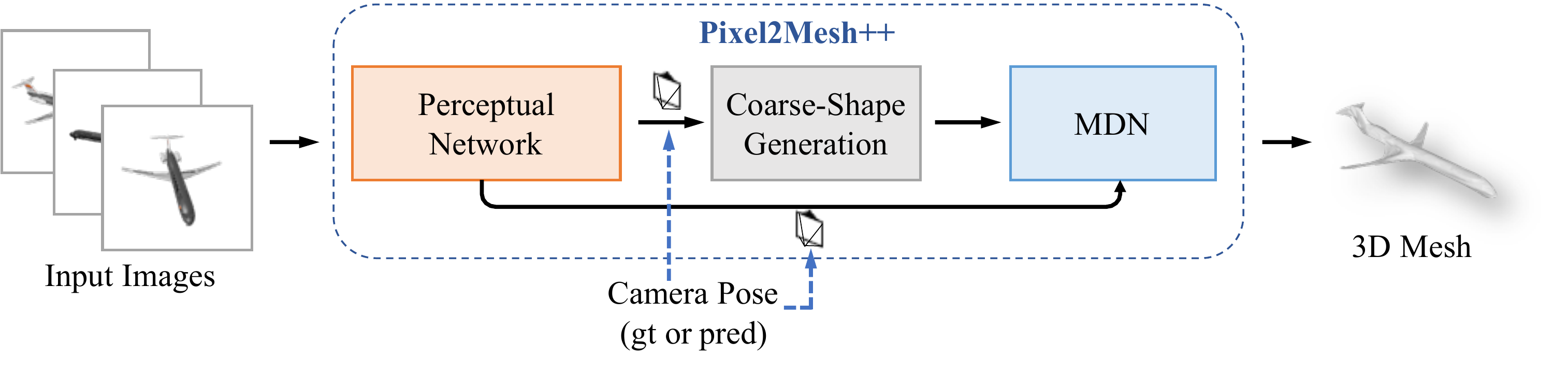}
	\caption{{\bf Overview of our framework.} Our method takes a small number of color images with ground truth or predicted camera poses as inputs and generates 3D meshes. The pretrained VGG-16 is used to extract perceptual features from input images. Details about the coarse shape generation and the MDN are shown in Fig.~\ref{fig:pipeline}.}
	\label{fig:pipe1}
\end{figure*}

\begin{figure*}[h!]
	\centering
	\includegraphics[width=\textwidth]{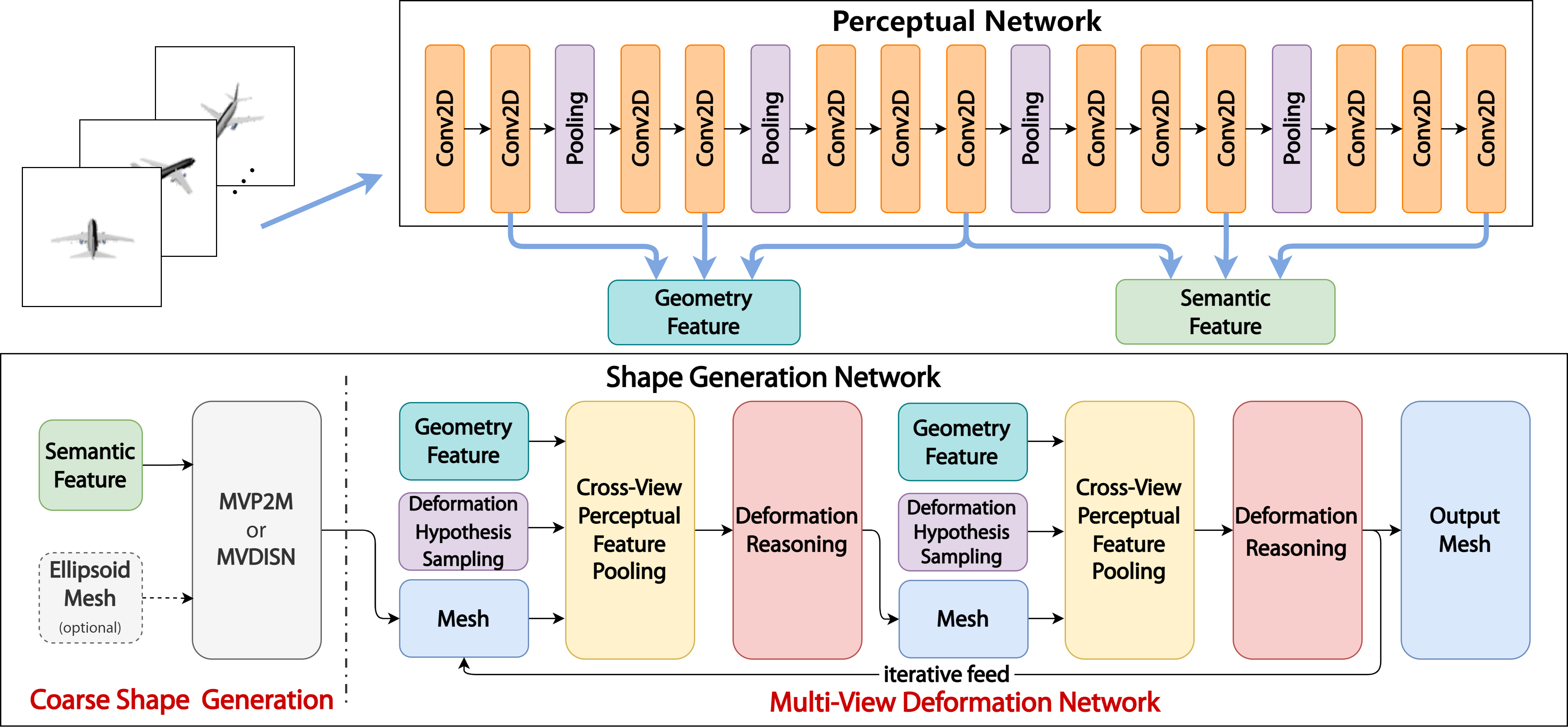}
	\caption{{\bf Shape Generation Pipeline.} Our 3D shape generation system consists of a 2D CNN extracting image features and a GCN deforming an ellipsoid to target shape. A coarse shape is generated from multi-view version Pixel2Mesh~\cite{wang2018pixel2mesh} or DISN~\cite{NIPS2019_DISN} and refined iteratively in Multi-View Deformation Network. To leverage cross-view information, our network pools perceptual features from multiple input images for hypothesis locations in the area around each vertex and predicts the optimal deformation.}
	\label{fig:pipeline}
\end{figure*}

Our model receives multiple color images of an object captured from different viewpoints (with or without known camera poses) and produces a 3D mesh model in the world coordinate. This procedure is conceptually visualized in Fig.~\ref{fig:pipe1}. 
Our approach consists of two parts: 3D shape prediction and camera pose estimation .
The 3D shape prediction framework adopts the strategy of coarse-to-fine (Fig. \ref{fig:pipeline}), in which a plausible but rough shape is generated first, and details are added later.
Realizing that existing 3D shape generators usually produce reasonable shape even from a single image, we simply use Pixel2Mesh \cite{wang2018pixel2mesh} or DISN \cite{NIPS2019_DISN} trained from multiple views to produce the coarse shape, which is taken as input to our Multi-View Deformation Network (MDN) for further improvement.
In MDN, each vertex first samples a set of deformation hypotheses from its surrounding area (Fig. \ref{fig:sampling} (a)). 
Each hypothesis then pools cross-view perceptual feature from early layers of a perceptual network , where the feature resolution is high and contains more low-level geometry information (Fig. \ref{fig:sampling} (b)).
These features are gathered by projection with the help of the ground truth or predicted camera parameters.
The cross-view perceptual features are further leveraged by the network to reason the best deformation to move the mesh vertex.
It is worth noting that our MDN can be applied iteratively for multiple times to gradually improve shapes.
In cases where we cannot obtain the camera parameters in advance, we estimate them through a neural network.
The camera pose estimation network predicts the compact 7D camera pose representation with respect to the canonical pose (Fig. \ref{fig:camnet}). 

\subsection{Multi-View Deformation Network}
In this section, we introduce Multi-View Deformation Network, which is the core of our system to enable the network exploiting cross-view information for shape generation.
We assume that we know the extrinsics of the camera. (We will come back to address this issue in the next section.)
It first generates deformation hypotheses for each vertex and learns to reason an optimum using feature pooled from inputs.
Our model is essentially a GCN, and can be jointly trained with other GCN based models like Pixel2Mesh.
We refer reader to \cite{BronsteinBLSV17,KipfW16} for details about GCN, and Pixel2Mesh \cite{wang2018pixel2mesh} for graph residual block which will be used in our model.

\subsubsection{Deformation Hypothesis Sampling}
\begin{figure}[ht]
	\centering
	\includegraphics[width=\columnwidth]{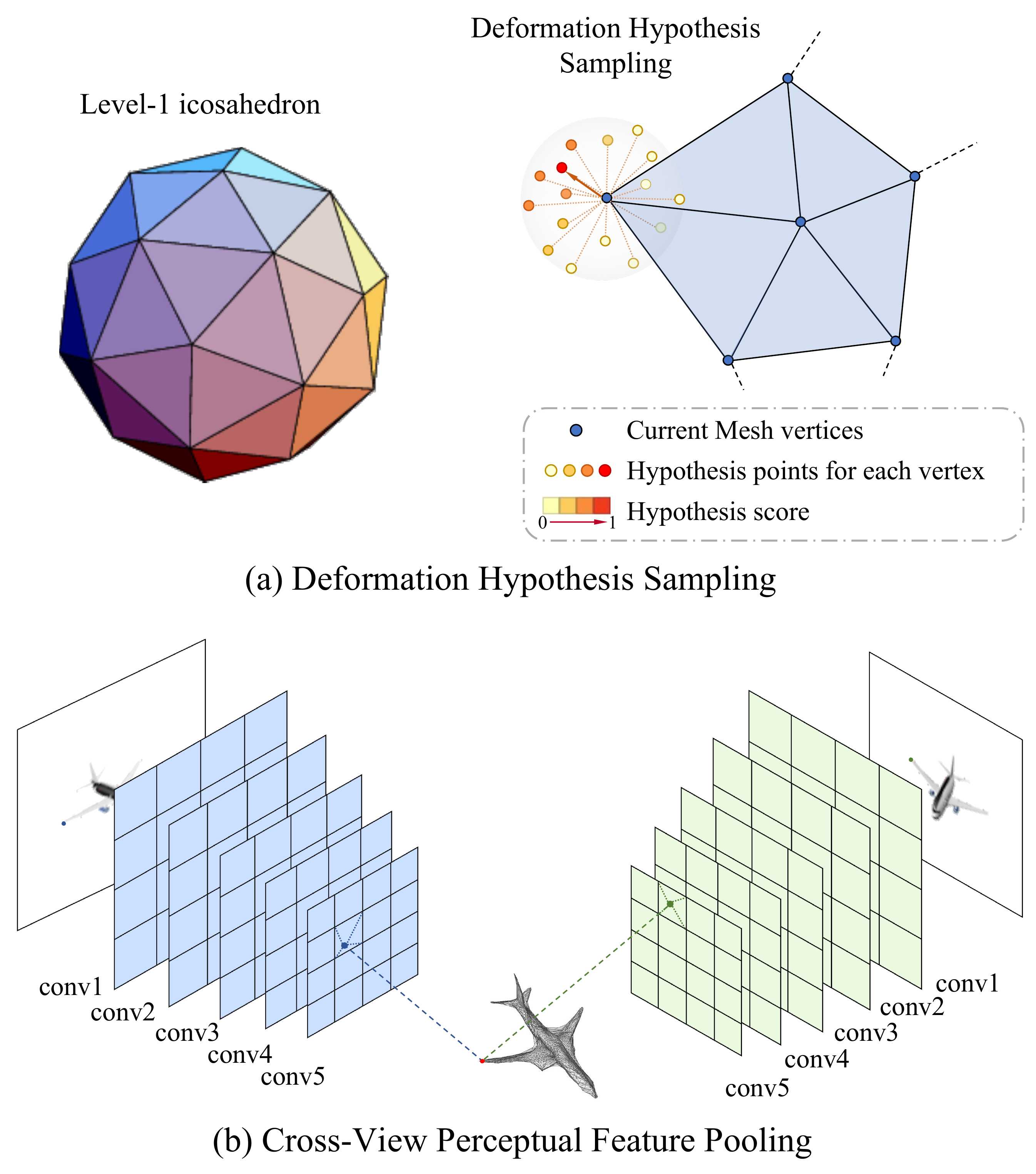}
	\caption{{\bf Deformation Hypothesis and Perceptual Feature Pooling.} (a) Deformation Hypothesis Sampling. We sample 42 deformation hypotheses from a level-1 icosahedron and build a GCN among hypotheses and the vertex. (b) Cross-View Perceptual Feature Pooling. The 3D vertex coordinates are projected to multiple 2D image planes using camera intrinsics and extrinsics. Perceptual features are pooled using bilinear interpolation, and feature statistics are kept on each hypothesis.}
	\label{fig:sampling}
\end{figure}

The first step is to propose deformation hypotheses for each vertex.
This is equivalent as sampling a set of target locations in 3D space where the vertex can be possibly moved to.
To uniformly explore the nearby area, we sample from a level-1 icosahedron centered on the vertex with a scale of 0.02, which results in 42 hypothesis positions (Fig. \ref{fig:sampling} (a), left).
We then build a local graph with edges on the icosahedron surface and additional edges between the hypotheses to the vertex in the center, which forms a graph with $43$ nodes and $120+42=162$ edges.
Such the local graph is built for all mesh vertices, and then fed into a GCN to predict vertex movements (Fig. \ref{fig:sampling} (a), right).

\subsubsection{Cross-View Perceptual Feature Pooling}\label{sec:cross}
The second step is to assign each node (in the local GCN) features from the multiple input color images.
Inspired by Pixel2Mesh, we use the prevalent VGG-16 architecture to extract perceptual features.
Since the camera poses are either as known inputs or can be predicted by our camera pose estimation network, each vertex and hypothesis can find their projections in all input color image planes using camera intrinsics and extrinsics and pool features from four neighboring feature grids using bilinear interpolation (Fig. \ref{fig:sampling} (b)).
Different from Pixel2Mesh where high level features from later layers of the VGG (i.e. `conv3\_3', `conv4\_3’, and `conv5\_3’) are pooled to better learn shape priors, MDN pools features from early layers (i.e. `conv1\_2', `conv2\_2’, and `conv3\_3’), which are in high spatial resolution and considered maintaining more detailed information.

To combine multiple features, concatenation has been widely used as a loss-less way;  however, it ends up with total dimension changing with respect to the number of input images.
Statistics feature has been proposed for multi-view shape recognition \cite{su15mvcnn} to tackle this problem.
Inspired by this work, we concatenate some statistics ($mean$, $max$, and $std$) of the features pooled from all views for each vertex, which makes our network naturally adaptive to variable input views and behave invariant to different input orders.
This also encourages the network to learn from cross-view feature correlations rather than each individual feature vector.
In addition to image features, we also concatenate the 3-dimensional vertex coordinate into the feature vector. 
In total, we compute for each vertex and hypothesis a 339 dimension feature vector.

\begin{figure*}[t]
	\centering
	\includegraphics[width=\textwidth]{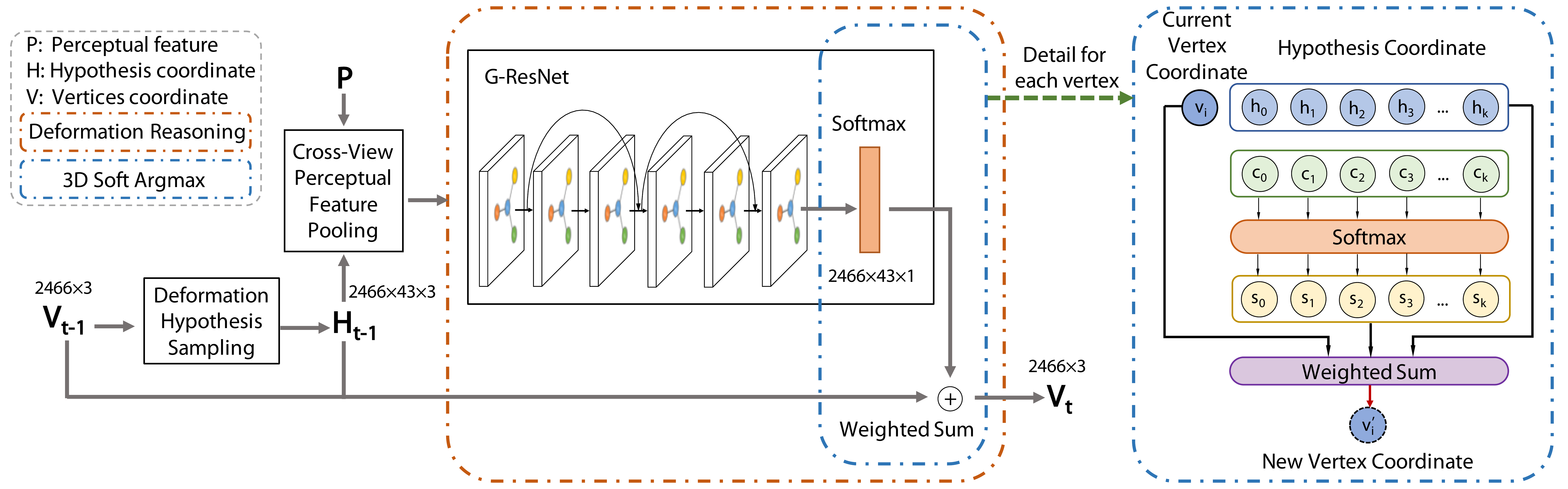}
	\caption{{\bf Deformation Reasoning.} The goal is to reason a good deformation from the hypotheses and pooled features. We first estimate a weight (green circle) for each hypothesis using a GCN. The weights are normalized  by a softmax layer (yellow circle), and the output deformation is the weighted sum of all the deformation hypotheses.}
	\label{fig:ld}
\end{figure*}

\subsubsection{Deformation Reasoning}
The next step is to reason an optimal deformation for each vertex from the hypotheses using pooled cross-view perceptual features.
Note that picking the best hypothesis of all needs an argmax operation, which requires stochastic optimization and usually is not optimal.
Instead, we design a differentiable network component to produce desirable deformation through soft-argmax of the 3D deformation hypotheses, which is illustrated in Fig. \ref{fig:ld}.
Specifically, we first feed the cross-view perceptual feature $P$ into a scoring network, consisting of 6 graph residual convolution layers \cite{wang2018pixel2mesh} activated by ReLU, to predict a scalar weight $c_i$ for each hypothesis.
All the weights are then fed into a softmax layer and normalized to scores $s_i$, with $\sum_{i=1}^{43} s_i=1$.
The vertex location is then updated as the weighted sum of all the hypotheses,
i.e. $v=\sum_{i=1}^{43} s_i*h_i$, where $h_i$ is the location of each deformation hypothesis including the vertex itself.
This deformation reasoning unit runs on all local GCN built upon every vertex with shared weights, as we expect all the vertices leveraging multi-view feature in a similar fashion.

\subsubsection{Differentiable Renderer}
Lastly, as an option, our method can be combined with differentiable renderer to achieve instance optimization during inference. 
We use common mesh rasterizer renderer pipeline~\cite{liu2020general,ravi2020pytorch3d} as an optional component.
When the silhouettes of images can be obtained from the input object, the 3D shape of the object can be further optimized by matching the silhouette of the predicted shapes with the multi-view input silhouette images through differentiable renderer.

\begin{figure}[ht]
	\centering
	\includegraphics[width=\columnwidth]{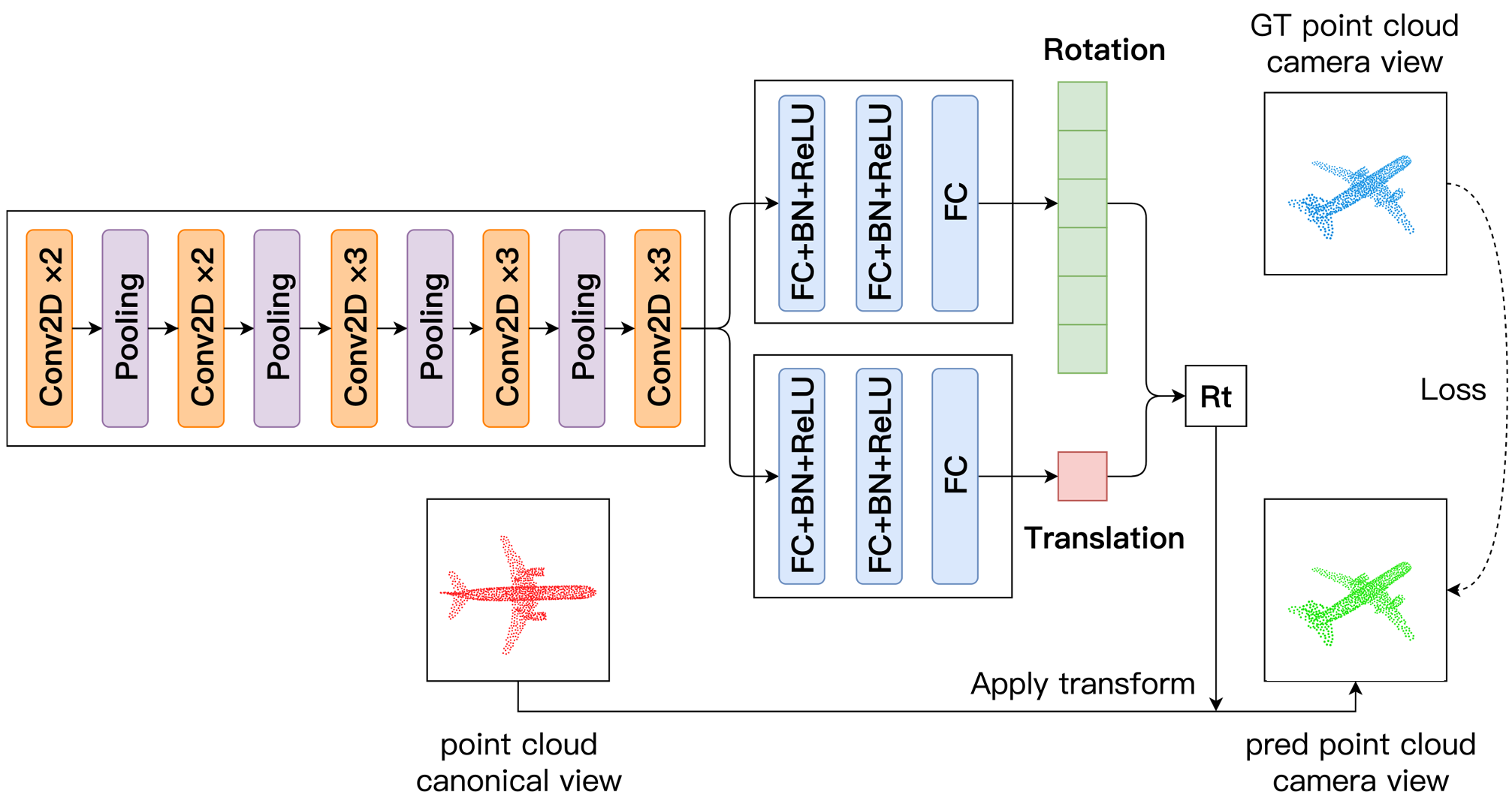}
	\caption{{\bf Network Architecture of Camera Pose Estimation Network.} Our camera pose estimation network predicts transformation from canonical view to camera view.
	The rotation is represented as a continuous 6D vector~\cite{zhou2019continuity} and the translation is defined as 1D distance between object center and camera.
	The predicted camera poses and the ground truth value are applied to the canonical view point cloud respectively to calculate the loss.}
	\label{fig:camnet}
\end{figure}

\subsection{Camera Pose Estimation Network}
In this section, we introduce camera pose estimation network.
In case the poses of the camera cannot be obtained, we use another convolution network to estimate them.

Given an input image without camera pose, our goal is to estimate the corresponding camera extrinsics. The models in the ShapeNet Core dataset~\cite{chang2015shapenet} are aligned for each category. Therefore, the world space is aligned with respect to this canonical coordinate system. As discussed in~\cite{insafutdinov2018unsupervised}, regressing camera pose directly often fails, Zhou \etal~\cite{zhou2019continuity} proposed a continuous rotation representation. We inherit this representation and regard the camera translation as the offset of the object along the direction of the camera optical axis, so that the camera extrinsics can be represented as 6D rotation and 1D translation. As illustrated in Fig.~\ref{fig:camnet}, we use a simple convolution network to regress the rotation and translation. The network first extract image feature through convolution layers, then using two branches to predict 6D rotation and 1D translation 
separately; and each branch  consists of multiple fully connected layers.

\subsection{Loss}
\noindent{\textbf{3D Shape Generation.}}
We train our model fully supervised using ground truth 3D CAD models.
Our loss function includes all terms from Pixel2Mesh, but extends the Chamfer distance loss to a re-sampled version.
Chamfer distance measures ``distance'' between two point clouds, which can be problematic when points are not uniformly distributed on the surface.
We propose to randomly re-sample the predicted mesh when calculating Chamfer loss using the re-parameterization trick proposed in \cite{ladicky2017point,smith19a}.
Specifically, given a triangle defined by 3 vertices $\left\{ v _ { 1 } , v _ { 2 } , v _ { 3 } \right\} \in \mathbb { R } ^ { 3 }$, a uniform sampling can be achieved by:
$$
s = \left( 1 - \sqrt { r _ { 1 } } \right) v _ { 1 } + \left( 1 - r _ { 2 } \right) \sqrt { r _ { 1 } } v _ { 2 } + \sqrt { r _ { 1 } } r _ { 2 } v _ { 1 },
$$
where $s$ is a point inside the triangle, and $r _ { 1 } , r _ { 2 } \sim U [ 0,1 ]$.
Knowing this, when calculating the loss, we uniformly sample our generated mesh for 4000 points, with the number of points per triangle proportional to its area.
We find this is empirically sufficient to produce a uniform sampling on our output mesh with 2466 vertices, and calculating Chamfer loss on the re-sampled point cloud, containing 6466 in total, helps to remove artifacts in the results.
The overall learning objective is as follows,
$$\mathcal{L} = \lambda_1 \mathcal{L}_{chamfer} + \lambda_2 \mathcal{L}_{normal} + \lambda_3  \mathcal{L}_{edge} + \lambda_4 \mathcal{L}_{lap}, $$
where $\mathcal{L}_{chamfer}$, $\mathcal{L}_{normal}$, $\mathcal{L}_{edge}$ and $\mathcal{L}_{lap}$ denotes  re-sampled chamfer loss, normal loss, edge length regularization and Laplacian regularization. Loss weights $\lambda_1=1$, $\lambda_2=1.6\times10^{-4}$, $\lambda_3=0.1$ and $\lambda_4=0.5$ are the hyperparameters which help to balance the losses. 
Moreover, we use an extra $L_2$ silhouette loss by comparing the projected silhouette between predicted shape and ground truth when using the differentiable renderer during inference.

\noindent{\textbf{Camera Pose Estimation.}} Following Xu~\etal ~\cite{NIPS2019_DISN}, we calculate $L_2$ distance between the ground truth point cloud in the camera view and the point cloud transformed from the canonical view by the predicted camera parameters. Specifically, the loss function is as follows:
\begin{gather*}
	\mathcal{L}_{cam} = \left\|\mathcal{T}^{\prime}(p) - \mathcal{T}(p) \right\|_{2}^{2} ,\\
	\mathcal{T}^{\prime}(p) = \mathbf{R}^{\prime} p + \mathbf{t}^{\prime} ,\\
	\mathcal{T}(p) = \mathbf{R} p + \mathbf{t} ,
\end{gather*}
where $p$ denotes point cloud in canonical view, $\mathcal{T}^{\prime}(p)$ and $\mathcal{T}(p)$ denotes the predict and ground truth transformation respectively, $\mathbf{R}$ denotes rotation and $\mathbf{t}$ denotes translation, and the symbol with superscript ``~$^{\prime}$~'' indicates the estimated value. Both $\mathcal{T}$ and $\mathcal{T}^{\prime}$ transform point cloud from canonical coordinates  to camera coordinates.

\subsection{Variants of Pixel2mesh++}

As in Fig.~\ref{fig:pipe1}, we introduce two variants of Pixel2mesh++ in term of different coarse-shape generation subnetwork.

\noindent \textbf{Na\"ive Methods for multi-view shape generation.} By directly extending previous works (Choy \etal \cite{choy20163d} and Kar \etal \cite{kar2017learning} )  to multiple input images, there are several n\"aive ways of shape generation from multi-view images, as compared in Tab.~\ref{tb1:multiview}. Essentially, these methods produce the coarse meshes, which can be utilized as the initial coarse mesh module in our Pixel2mesh++. Specifically, \\
(1) \textit{P2M-M}: we directly run single-view Pixel2Mesh on each of the input image and fuse multiple results \cite{curless1996volumetric,LorensenC87}. \\
(2) \textit{MVP2M}: we extend Pixel2Mesh to access multiple images in a single network forward pass by having it pools multi-view features from all the inputs. This can be achieved by replacing the perceptual feature pooling layers with our multi-view version as introduced in Sec. \ref{sec:cross}. In particular, perceptual features are pooled from layer `conv3\_3', `conv4\_3', and `conv5\_3' from the VGG-16 network, and feature statistics (Sec. \ref{sec:stat_feat}) are calculated and concatenated, which ends up with a 1280 dimension feature vector. 
In practice, we also tried to pool geometry related features from early convolution layers (i.e. `conv1\_2', `conv2\_2', and `conv3\_3'), but found it doesn't work as well as the case with semantic feature pooled from later layers.
Using a template with a fixed topology will limit the expression ability of the shape. \\
(3) \textit{MVDISN}:  in order to show the generalization of our method in arbitrary topology, we build another implicit representation baseline based on DISN. We equip DISN with our perceptual feature pooling layers and extend it to a multi-view version (Tab.~\ref{tb1:multiview}, MVDISN). The output signed distance function of MVDISN is converted to mesh using marching cubes algorithm.

\noindent \textbf{Our Variants.} We thus define two variants of our Pixel2mesh++ via using different initial coarse meshes, which are produced by MVP2M and MVDISN.
Particularly, we have the variants as follows,\\
(1) \textit{Ours-P}: As in Fig.~\ref{fig:pipe1}, we take MVP2M as the initial coarse shape, and combine with our MDN. This variant needs ellipsoid mesh as input for coarse shape generation. We train the model in an end-to-end manner. \\
(2) \textit{Ours-D}: Since using a template with a fixed topology will limit the expression ability of the shape. We combine the implicit methods MVDISN with stronger representation ability as initialization of the coarse shape. The variant of Ours-D is also learned end-to-end in our experiments. Note that MVDISN can also produce the coarse meshes of objects of arbitrary topology, and our method can still efficiently improve the quality of object meshes.

\section{Experiments}
In this section, we perform extensive evaluation of our model for multi-view shape generation.
We compare to state-of-the-art methods, as well as conduct controlled experiments with respect to (\wrt) various aspects, \eg cross category generalization, quantity of inputs, etc. All the experiments assume camera poses are known without specification.

\subsection{Experimental setup}
\subsubsection{Dataset}
We adopt the dataset provided by Choy \etal \cite{choy20163d} as it is widely used by many existing 3D shape generation works. The dataset is created using a subset of ShapeNet\cite{chang2015shapenet} containing 50k 3D CAD models from 13 categories. Each model is rendered from 24 randomly chosen camera viewpoints, and the camera intrinsic and extrinsic parameters are given. For fair comparison, we use the same training/testing split as in Choy \etal \cite{choy20163d} for all our experiments. The ground truth models are transformed to camera coordinate based on the camera parameters from the Choy \etal  \cite{choy20163d}, and scaled by a factor of 0.57 to aligned with rendering images following Wang \etal \cite{wang2018pixel2mesh}.
\subsubsection{Evaluation Metric}
\noindent{\textbf{3D Shape Generation.}}
As for 3D shape, we use standard evaluation metrics for 3D shape generation. Following Fan \etal \cite{fan2017point}, we calculate Chamfer Distance(CD) between 2048 points uniformly sampled from the ground truth and our prediction to measure the surface accuracy. We also use F-score following Wang \etal \cite{wang2018pixel2mesh} to measure the completeness and precision of generated shapes\textcolor{black}{, where $\tau$=1.0$\times10^{-4}$}. For CD, the smaller is better. For F-score, the larger is better. \textcolor{black}{We also evaluate volumetric IoU. We transform the ground truth and predictions back to canonical view to calculate IoU.
We obtain unbiased estimates of the volumetric IoU by randomly sampling 100k points following Mescheder~\etal~\cite{mescheder2019occupancy}.}

\noindent{\textbf{Camera Pose Estimation.}}
We use four types of standard metrics to evaluate our camera pose estimation method: 2D reprojection error $d_{2D}$, mean distance $d_{3D}$~\cite{NIPS2019_DISN}, 2D reprojection accuracy Acc$_{2D}$ and average 3D distance of model points (ADD) accuracy metric ADD$_{3D}$~\cite{peng2019pvnet}.
The reprojection error $d_{2D}$ is compared with the image size $137\times 137$ pixels. Different from the threshold selected for general 6 DOF estimation tasks~\cite{peng2019pvnet}, we use more strict criteria. Specifically, we set the threshold to 2 for the Acc$_{2D}$ metric and the threshold to $5\%$ for the ADD$_{3D}$ metric.

\subsubsection{Implementation Details}
\noindent{\textbf{3D Shape Generation.}}
For initialization, we use Pixel2Mesh~\cite{wang2018pixel2mesh} to generate a coarse shape with 2466 vertices, or use DISN~\cite{NIPS2019_DISN} to generate a coarse shape with varying topology.
To improve the quality of initial mesh, we equip these methods with our cross-view perceptual feature pooling layer, which allows it to extract features from multiple views.
We denote these multi-view version shape generation methods as MVP2M and MVDISN.
The network is implemented in Pytorch and optimized using Adam with weight decay as $5\times 10^{-6}$ and mini-batch size as 1. 
The model use MVP2M as coarse shape generation method is trained for 40 epochs with learning rate $1\times 10^{-5}$ over 72 hours, and the model use MVDISN is trained for 100 epochs with learning rate $3 \times 10^{-4}$ over 96 hours. The MDN network is trained for another 20 epoch over 24 hours with the learning rate as $1\times 10^{-6}$.
The overall network needs to learn the statistical information between the multi-view features, and the MDN module needs plausible initialization from the coarse shape generation stage to effectively refine the 3D shape. Therefore, the convergence speed of our method is slower than single view deep models and requires longer training time.
During training, we randomly pick three images for a mesh as input. 
All models are trained on a NVIDIA Titan Xp GPU.
During test, it takes 0.32s to produce a mesh.

\noindent{\textbf{Camera Pose Estimation.}}
The camera parameter estimation network is trained for 70 epochs, over 48 hours. The learning rate is set to $1\times 10^{-4}$. All 24 views of a mesh and the ground truth mesh points near surface are used as training data. During inference, we fed a single image to the network to obtain the corresponding camera pose.

\begin{table*}[h]
  \centering
  \caption{{\bf Comparison to Multi-view Shape Generation Methods.} We show F-score on each semantic category. Our model significantly outperforms previous methods, i.e. 3DR2N2 \cite{choy20163d} and LSM \cite{kar2017learning}, and competitive baselines derived from Pixel2Mesh \cite{wang2018pixel2mesh}. The notation $\dagger$ indicates the methods which does not require camera extrinsics.}
  \begin{tabu} to \textwidth {X[1.5,l]X[1.5,c]X[c]X[1.5,c]X[1.5,c]X[1.5,c]X[1.5c]X[1.5c]X[1.5,c]X[1.2,c]X[1.5,c]X[1.5,c]X[1.5,c]X[1.5c]X[1.5c]}
    \toprule
    \multirow{3}{*}{Category} &
    \multicolumn{7}{c}{F-score($\tau$) $\uparrow$}&
    \multicolumn{7}{c}{F-score($2\tau$) $\uparrow$}\\
    \tabuphantomline
    \cmidrule(lr){2-8} \cmidrule(lr){9-15}
    & {\scriptsize{}3DR2N2$^{\dagger}$} & {\scriptsize{}LSM} & {\scriptsize{}P2M-M} & {\scriptsize{}MVP2M} & {\scriptsize{}Ours-P} & {\scriptsize{}MVDISN} & {\scriptsize{}Ours-D} & {\scriptsize{}3DR2N2$^{\dagger}$} & {\scriptsize{}LSM} & {\scriptsize{}P2M-M} & {\scriptsize{}MVP2M} & {\scriptsize{}Ours-P} & {\scriptsize{}MVDISN} & {\scriptsize{}Ours-D} \\
    
    \midrule
    \midrule
    Couch & 45.47 & 43.02 & 53.70 & 52.53 & 57.53 & 68.61 & 71.66
          & 59.97 & 55.49 & 72.04 & 72.17 & 75.74 & 80.85 & 82.84\\
    Cabinet & 54.08 & 50.80 & 63.55 & 58.04 & 61.46 & 51.12 & 54.11
            & 64.42 & 60.72 & 79.93 & 77.28 & 79.46 & 67.71 & 69.72\\
    Bench & 44.56 & 49.33 & 61.14 & 60.20 & 67.70 & 74.22 & 80.14
          & 62.47 & 65.92 & 75.66 & 75.89 & 81.49 & 85.79 & 89.23\\
    Chair & 37.62 & 48.55 & 55.89 & 54.04 & 61.78 & 70.73 & 76.77
          & 54.26 & 64.95 & 72.36 & 72.04 & 77.99 & 83.00 & 86.72\\
    Monitor & 36.33 & 43.65 & 54.50 & 53.36 & 58.95 & 68.42 & 73.41
            & 48.65 & 56.33 & 70.51 & 70.34 & 75.03 & 82.97 & 85.89\\
    Firearm & 55.72 & 56.14 & 74.85 & 78.52 & 85.31 & 91.37 & 94.27
          & 76.79 & 73.89 & 84.82 & 88.67 & 92.64 & 96.43 & 97.75\\
    Speaker & 41.48 & 45.21 & 51.61 & 49.21 & 53.75 & 56.95 & 61.58
            & 52.29 & 56.65 & 68.53 & 67.66 & 71.20 & 70.30 & 74.02\\
    Lamp  & 32.25 & 45.58 & 51.00 & 50.46 & 63.40 & 65.89 & 74.39
          & 49.38 & 64.76 & 64.72 & 65.63 & 75.62 & 77.04 & 82.66\\
    Cellphone & 58.09 & 60.11 & 70.88 & 71.88 & 75.79 & 85.14 & 87.65
              & 69.66 & 71.39 & 84.09 & 85.24 & 87.72 & 92.58 & 94.31\\
    Plane & 47.81 & 55.60 & 72.36 & 76.67 & 84.17 & 81.37 & 85.72
          & 70.49 & 76.39 & 82.74 & 87.18 & 91.76 & 89.92 & 92.53\\
    Table & 48.78 & 48.61 & 67.89 & 66.35 & 70.52 & 67.59 & 71.78
          & 62.67 & 62.22 & 81.04 & 80.12 & 83.15 & 80.00 & 82.70\\
    Car   & 59.86 & 51.91 & 67.29 & 65.18 & 67.64 & 67.57 & 69.96
          & 78.31 & 68.20 & 84.39 & 82.38 & 83.62 & 79.48 & 80.82\\
    Watercraft & 40.72 & 47.96 & 57.72 & 59.21 & 65.98 & 71.10 & 76.62
               & 63.59 & 66.95 & 72.96 & 75.63 & 80.41 & 82.80 & 86.26\\
    \midrule
    Mean & 46.37 & 49.73 & 61.72 & 61.20 & \textbf{67.23} & 70.78 & \textbf{75.24}
         & 62.53 & 64.91 & 76.45 & 76.94 & \textbf{81.22} & 82.22 & \textbf{85.04} \\
    \bottomrule
  \end{tabu}
  \label{tb1:multiview}
\end{table*}

\begin{table*}[h]
  \centering
\caption{{\bf Comparison to Multi-view Shape Generation Methods.} We show the Chamfer Distance on each semantic category. Our method achieves the best performance overall. The notation $\dagger$ indicates the methods which does not require camera extrinsics.}
  \begin{tabu} to 0.7\textwidth {X[1.5,l]X[1.5,c]X[c]X[1.5,c]X[1.5,c]X[1,c]X[1.5,c]X[2,c]}
    \toprule
    \multirow{2}{*}{Category} &
    \multicolumn{7}{c}{Chamfer Distance(CD) $\downarrow$}\\
    \tabuphantomline
    \cmidrule(lr){2-8}
    & 3DR2N2$^{\dagger}$ & LSM & P2M-M & MVP2M & Ours-P & MVDISN & Ours-D\\
    \midrule
    \midrule
    Couch   & 0.806 & 0.730 & 0.496 & 0.440 & 0.383 & 0.396 & 0.362\\
    Cabinet & 0.613 & 0.634 & 0.359 & 0.381 & 0.354 & 0.765 & 0.737\\
    Bench   & 1.362 & 0.572 & 0.594 & 0.462 & 0.368 & 0.343 & 0.285\\
    Chair   & 1.534 & 0.495 & 0.561 & 0.494 & 0.402 & 0.409 & 0.346\\
    Monitor & 1.465 & 0.592 & 0.654 & 0.585 & 0.496 & 0.332 & 0.285\\
    Firearm & 0.432 & 0.385 & 0.428 & 0.250 & 0.184 & 0.165 & 0.137\\
    Speaker & 1.443 & 0.767 & 0.697 & 0.666 & 0.604 & 0.746 & 0.663\\
    Lamp    & 6.780 & 1.768 & 1.184 & 0.891 & 0.681 & 0.820 & 0.722\\
    Cellphone & 1.161 & 0.362 & 0.360 & 0.373 & 0.318 & 0.179 & 0.152\\
    Plane   & 0.854 & 0.496 & 0.457 & 0.251 & 0.183 & 0.271 & 0.226\\
    Table   & 1.243 & 0.994 & 0.441 & 0.463 & 0.406 & 0.587 & 0.532\\
    Car     & 0.358 & 0.326 & 0.264 & 0.260 & 0.243 & 0.362 & 0.344\\
    Watercraft & 0.869 & 0.509 & 0.627 & 0.413 & 0.337 & 0.332 & 0.279\\
    \midrule
    Mean & 1.455 & 0.664 & 0.548 & 0.456 & \textbf{0.381} & 0.439 & \textbf{0.390}\\
    \bottomrule
  \end{tabu}
  \label{tb2:multiviewcd}
\end{table*}

\begin{table}[h]
  \centering
  \caption{\textcolor{black}{{\bf Comparison to Baselines and Our Variants.} We show the volumetric IoU on each semantic category.}}
  \begin{tabu} to \columnwidth {X[1.5,l]X[1.5,c]X[1,c]X[1.5,c]X[1.5,c]}
    \toprule
    \multirow{2}{*}{Category} &
    \multicolumn{4}{c}{IoU $\uparrow$}\\
    \tabuphantomline
    \cmidrule(lr){2-5}
    & MVP2M & Ours-P & MVDISN & Ours-D\\
    \midrule
    \midrule
    Couch   & 0.546 & 0.569 & 0.671 & 0.688\\
    Cabinet & 0.536 & 0.550 & 0.398 & 0.397\\
    Bench   & 0.211 & 0.224 & 0.418 & 0.399\\
    Chair   & 0.343 & 0.362 & 0.515 & 0.509\\
    Monitor & 0.383 & 0.408 & 0.519 & 0.502\\
    Firearm & 0.371 & 0.426 & 0.621 & 0.539\\
    Speaker & 0.480 & 0.496 & 0.525 & 0.528\\
    Lamp    & 0.218 & 0.240 & 0.321 & 0.310\\
    Cellphone & 0.613 & 0.644 & 0.679 & 0.628\\
    Plane   & 0.424 & 0.484 & 0.556 & 0.533\\
    Table   & 0.306 & 0.314 & 0.453 & 0.442\\
    Car     & 0.513 & 0.523 & 0.602 & 0.596\\
    Watercraft & 0.391 & 0.429 & 0.565 & 0.556\\
    \midrule
    Mean & 0.411 & 0.436 & 0.527 & 0.508\\
    \bottomrule
  \end{tabu}
  \vspace{1mm}
  \label{tbl:iou}
\end{table}

\subsection{Comparison to Multi-view Shape Generation}
\label{sec:mvp2m}

Tab.~\ref{tb1:multiview} shows quantitative comparison in F-score. 
As can be seen, our baselines already outperform other methods, which show the advantage of mesh representation in finding surface details.
Moreover, directly equipping Pixel2Mesh with multi-view features does not improve too much (even slightly worse than the average of multiple runs of single-view Pixel2Mesh), which shows dedicate architecture is required to efficiently learn from multi-view features.
In contrast, our Multi-View Deformation Network significantly further improves the results from the MVP2M baseline (i.e. our coarse shape initialization).
It is worth noting that the MVDISN method has more accurate initial shapes and is a stronger baseline. Therefore, using this baseline directly is better than using MVP2M. However, by combining it with our MDN method, the performance will also be significantly improved.

Tab.~\ref{tb2:multiviewcd} shows quantitative comparison in Chamfer Distance. Our method is not only better than other multi-view shape generation methods, but also has a consistent performance improvement for the baseline algorithm we constructed.

\textcolor{black}{
We also compare quantitative results for volumetric IoU. The calculation of IoU may be affected by self-intersecting faces and non-watertight surfaces, since it is complicated to determine whether a point is inside self-intersecting faces.
As shown in Tab.~\ref{tbl:iou}, when the initial coarse shape generated from the ellipsoid, Ours-P has an improvement compared to MVP2M. However, when the topology changes, Ours-D has a slightly lower IoU than MVDISN due to the complicated procedures to determine the inside points.
}

\begin{figure*}[th]
	\centering
	\includegraphics[width=\textwidth]{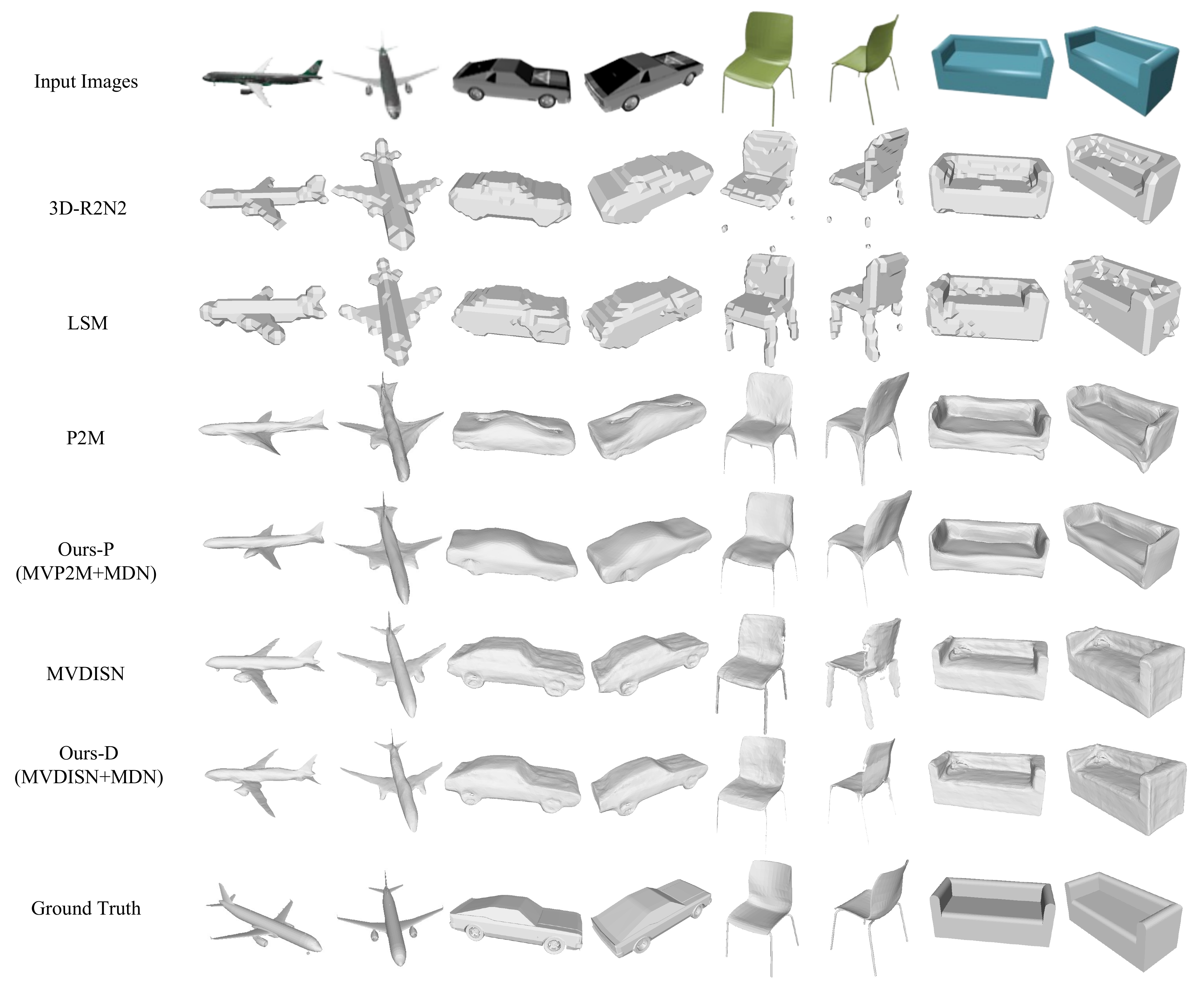}
	\caption{{\bf Qualitative Evaluation.} From top to bottom, we show in each row: two camera views, results of 3DR2N2, LSM, multi-view Pixel2Mesh, ours, and the ground truth. Our predicts maintain good details and align well with different camera views.
	Please see supplementary materials for more results. 
	}
	\label{fig:multiview}
\end{figure*}

More qualitative results are shown in Fig. \ref{fig:multiview}.
We show results from different methods aligned with one input view (left) and a random view (right).
As can be seen, Choy \etal \cite{choy20163d} (3D-R2N2) and Kar \etal \cite{kar2017learning} (LSM) produce 3D volume, which lose thin structures and surface details.
Pixel2Mesh (P2M) produces mesh models but shows obvious artifacts when visualized in viewpoint other than the input.
In comparison, our results contain better surface details and more accurate geometry learned from multiple views.

\subsection{Generalization Capability}
Our MDN is inspired by multi-view geometry methods, where 3D location is reasoned via cross-view information.
In this section, we investigate the generalization capability of MDN in different aspects to improve the initialization mesh. For all the experiments in this section, we fix the coarse stage and train/test MDN under different settings.

\begin{figure}
\scalebox{0.8}{
    \centering{}%
    \begin{tabular}{cc}
    \hspace{-4mm}
    \begin{tabular}{c}
    {\footnotesize{}}%
    \renewcommand\tabcolsep{1.8pt}
    \begin{tabular}{c|c|c|c}
    \toprule 
    \multirow{2}{*}{{\footnotesize{}Category}} & \multicolumn{3}{c}{\textcolor{black}{\footnotesize{}F-score($\tau$) $\uparrow$}}\tabularnewline
    \cmidrule{2-4} \cmidrule{3-4} \cmidrule{4-4} 
     & {\footnotesize{}Corase} & {\footnotesize{}Except} & {\footnotesize{}All}\tabularnewline
    \midrule
    \midrule 
    {\footnotesize{}lamp} & {\footnotesize{}50.82} & {\footnotesize{}61.78}\textcolor{brown}{\footnotesize{}(+10.9)} & {\footnotesize{}62.55}\textcolor{brown}{\footnotesize{}(+11.7)}\tabularnewline
    {\footnotesize{}cabinet} & {\footnotesize{}56.85} & {\footnotesize{}65.73}\textcolor{brown}{\footnotesize{}(+8.88)} & {\footnotesize{}65.72}\textcolor{brown}{\footnotesize{}(+8.99)}\tabularnewline
    {\footnotesize{}cellphone} & {\footnotesize{}66.07} & {\footnotesize{}73.17}\textcolor{brown}{\footnotesize{}(+7.10)} & {\footnotesize{}74.36}\textcolor{brown}{\footnotesize{}(+8.29)}\tabularnewline
    {\footnotesize{}chair} & {\footnotesize{}54.19} & {\footnotesize{}60.68}\textcolor{brown}{\footnotesize{}(+6.49)} & {\footnotesize{}62.05}\textcolor{brown}{\footnotesize{}(+7.86)}\tabularnewline
    {\footnotesize{}monitor} & {\footnotesize{}53.41} & {\footnotesize{}59.47}\textcolor{brown}{\footnotesize{}(+6.06)} & {\footnotesize{}60.01}\textcolor{brown}{\footnotesize{}(+6.60)}\tabularnewline
    {\footnotesize{}speaker} & {\footnotesize{}48.90} & {\footnotesize{}54.65}\textcolor{brown}{\footnotesize{}(+5.75)} & {\footnotesize{}54.88}\textcolor{brown}{\footnotesize{}(+5.98)}\tabularnewline
    {\footnotesize{}table} & {\footnotesize{}65.95} & {\footnotesize{}71.39}\textcolor{brown}{\footnotesize{}(+5.44)} & {\footnotesize{}71.89}\textcolor{brown}{\footnotesize{}(+5.94)}\tabularnewline
    {\footnotesize{}bench} & {\footnotesize{}60.37} & {\footnotesize{}65.67}\textcolor{brown}{\footnotesize{}(+5.30)} & {\footnotesize{}66.24}\textcolor{brown}{\footnotesize{}(+5.87)}\tabularnewline
    {\footnotesize{}couch} & {\footnotesize{}53.17} & {\footnotesize{}56.93}\textcolor{brown}{\footnotesize{}(+3.76)} & {\footnotesize{}57.56}\textcolor{brown}{\footnotesize{}(+4.39)}\tabularnewline
    {\footnotesize{}plane} & {\footnotesize{}75.16} & {\footnotesize{}76.28}\textcolor{brown}{\footnotesize{}(+1.12)} & {\footnotesize{}76.79}\textcolor{brown}{\footnotesize{}(+1.63)}\tabularnewline
    {\footnotesize{}firearm} & {\footnotesize{}79.66} & {\footnotesize{}80.33}\textcolor{brown}{\footnotesize{}(+0.67)} & {\footnotesize{}80.73}\textcolor{brown}{\footnotesize{}(+1.07)}\tabularnewline
    {\footnotesize{}watercraft} & {\footnotesize{}61.85} & {\footnotesize{}62.06}\textcolor{brown}{\footnotesize{}(+0.21)} & {\footnotesize{}62.99}\textcolor{brown}{\footnotesize{}(+1.14)}\tabularnewline
    {\footnotesize{}car} & {\footnotesize{}67.26} & {\footnotesize{}67.40}\textcolor{brown}{\footnotesize{}(+0.14)} & {\footnotesize{}68.44}\textcolor{brown}{\footnotesize{}(+1.18)}\tabularnewline
    \bottomrule
    \end{tabular}\tabularnewline
    \end{tabular} &
    \hspace{-0.35in}%
    \begin{tabular}{c}
    \includegraphics[width=0.3\textwidth]{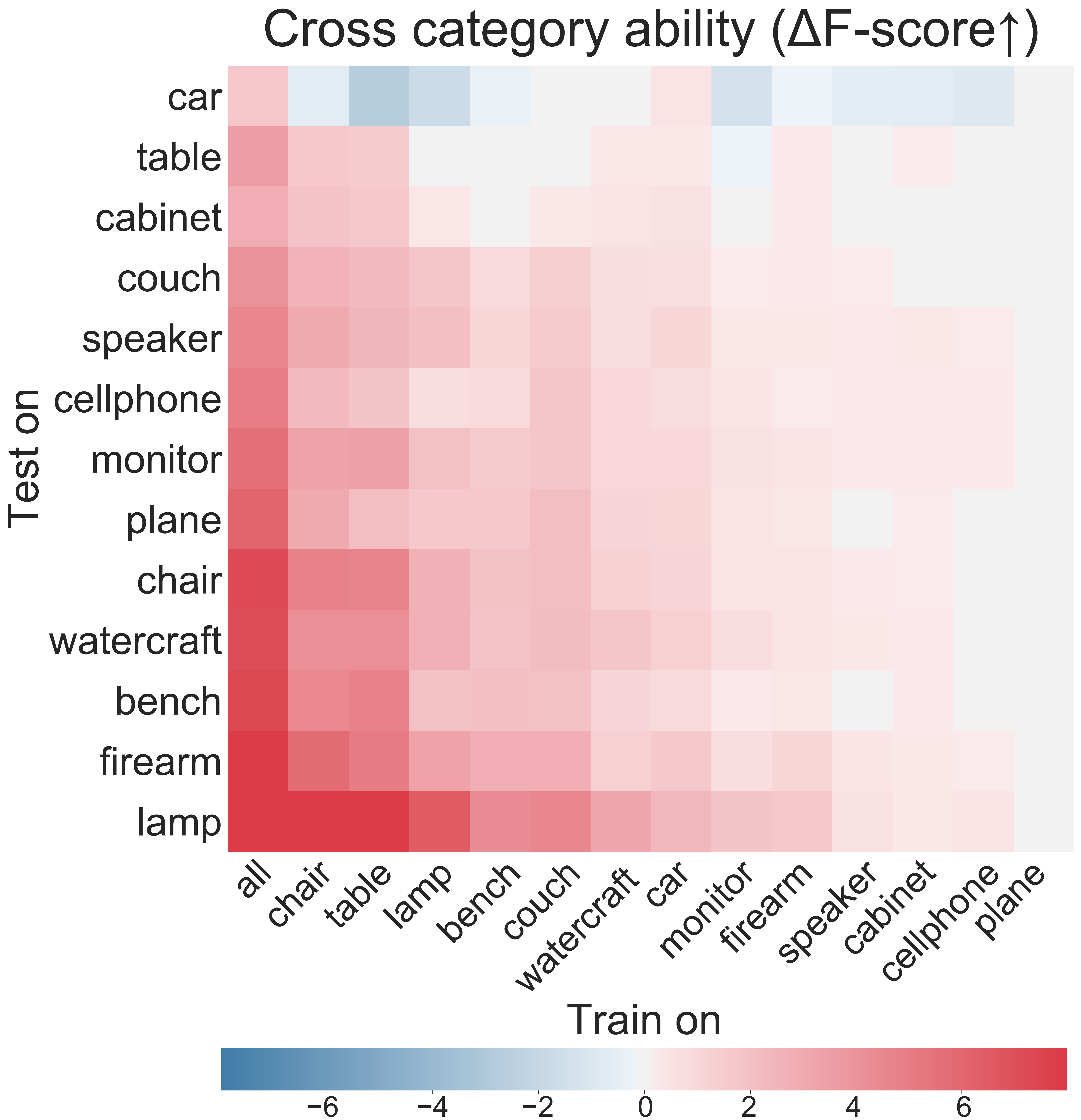}\tabularnewline
    \end{tabular} 
    \tabularnewline
    \makecell{\small{}(a) Absolute and relative improvements\\ of training except one category} & \makecell{\small{}(b) Relative improvements of training\\ with one category}\tabularnewline
    \end{tabular}
}
    \vspace{1mm}
    \caption{\textcolor{black}{{\bf Quantitative Results of Cross-Category Generalization.} (a) Improved F-score($\tau$) and absolute F-score($\tau$) of MDN trained on 12 out of 13 categories and tested on the one left. (b) MDN trained on 1 category and tested on the others. Each block represents the improved F-score($\tau$) of MDN trained on horizontal category and tested on vertical category.}}
    \label{fig:cross}
\end{figure}

\begin{figure}[tb]
 \centering 
 \includegraphics[width=\linewidth]{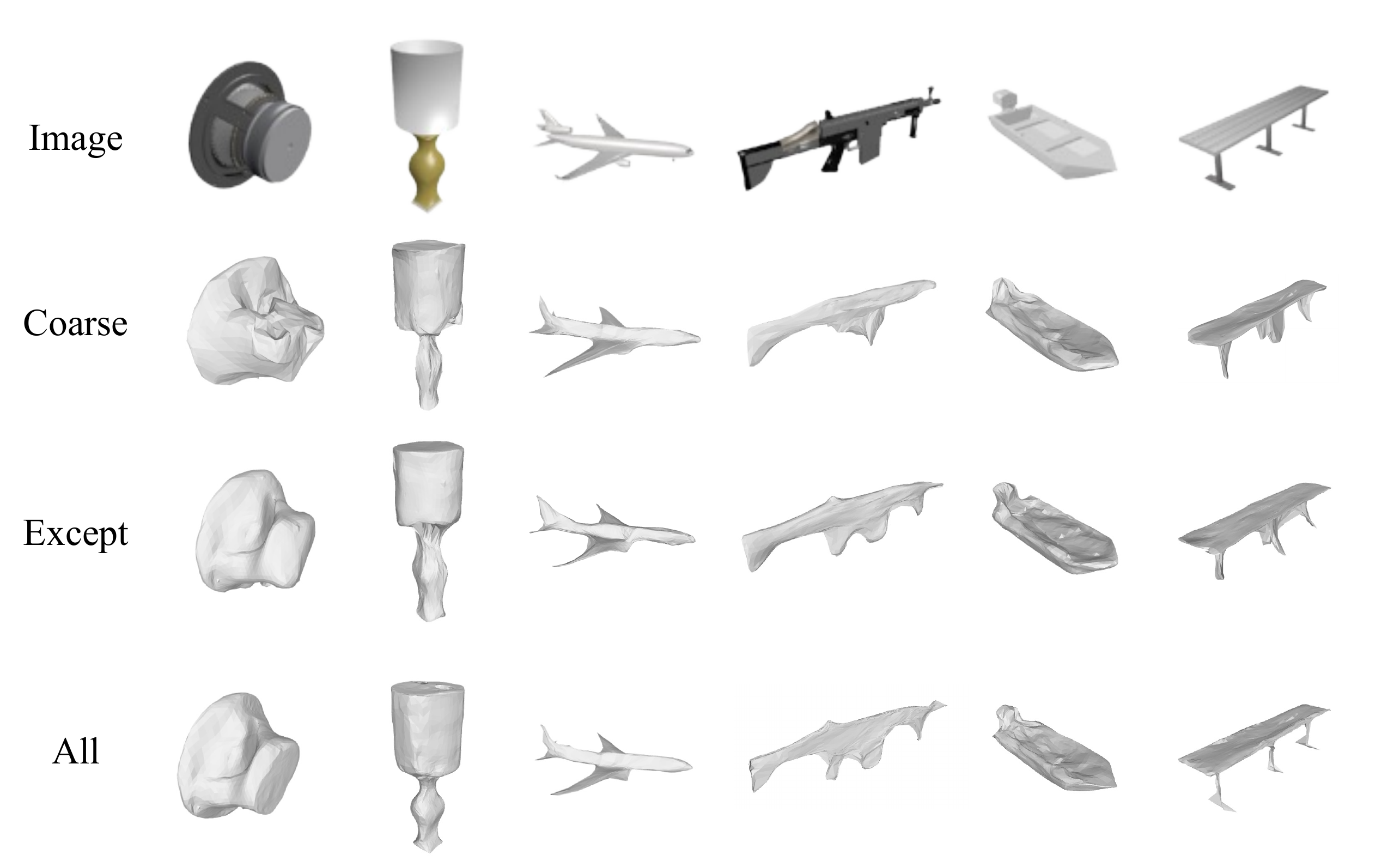}
 \caption{\textcolor{black}{{\bf Qualitative Results of Cross-Category Generalization.} 
 `Coarse' means results from MVP2M. `Except' indicates that results are generated from the model that has not been trained with the same category. `All' indicates that the model has been trained with data from all categories.}}
 \label{fig:qualitative_cross}
\end{figure}

\begin{figure*}[t]
	\centering
	\includegraphics[width=\textwidth]{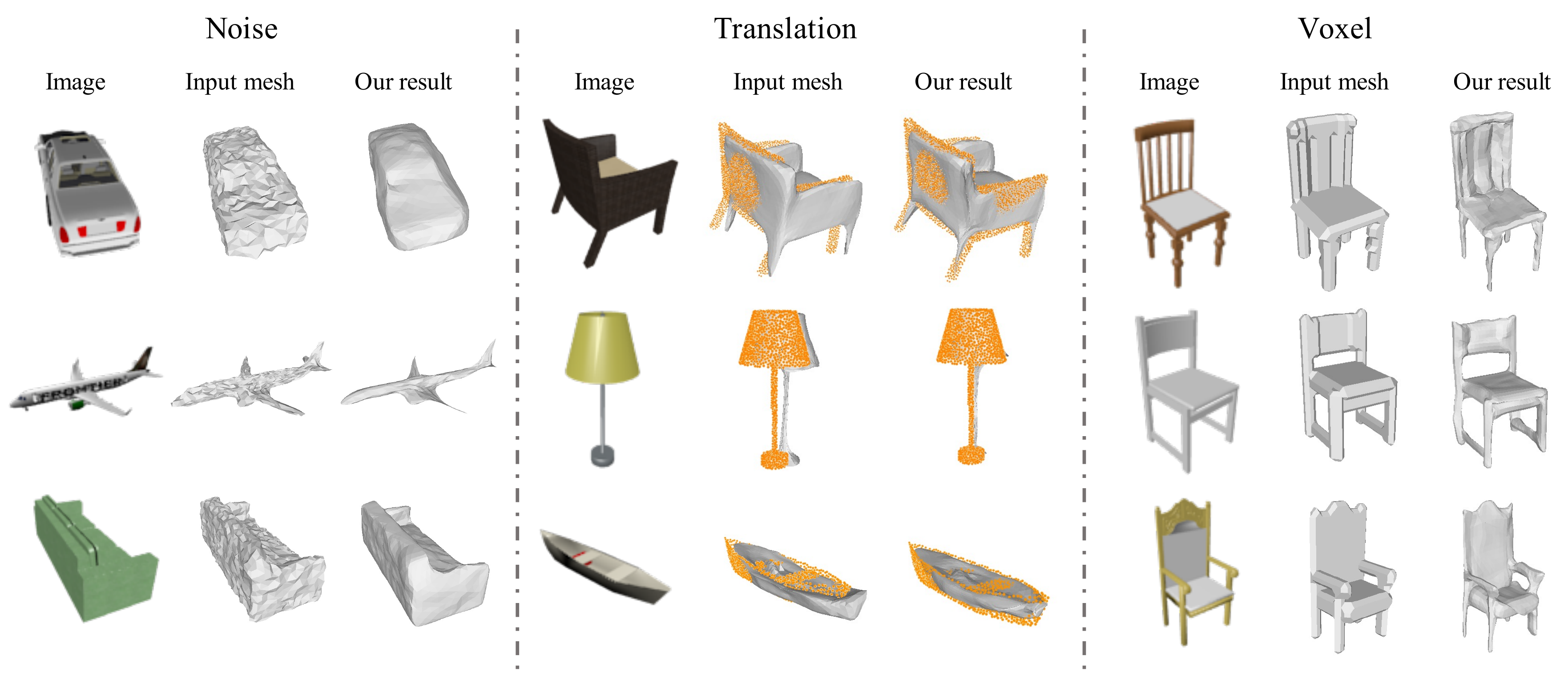}
	\caption{{\bf Robustness to Initialization.} Our model is robust to added noise, shift, and input mesh from other sources.}
	\label{fig:init}
\end{figure*}
\nopagebreak
\subsubsection{Semantic Category}
We first verify how our network generalizes across semantic categories.
We fix the initial MVP2M and train MDN with 12 out 13 categories and test on the one left out. \textcolor{black}{The improved and the absolute F-scores($\tau$) upon the initialization from Pixel2Mesh are both shown in Fig. \ref{fig:cross} (a).}
As can be seen, the performance is only slightly lower when the testing category is removed from the training set compared to the model trained using all categories.
To make it more challenging, we further train MDN on only one category and test on other categories.
Surprisingly, MDN still generalizes well among most categories as shown in Fig. \ref{fig:cross} (b). 
\textcolor{black}{Besides, we provide some qualitative results of the cross-category generalization in Fig.~\ref{fig:qualitative_cross}. These results show that our MDN enjoys good generalization for the unseen categories.}

Strong generalizing categories (\eg chair, table, lamp) tend to have relatively complex geometry, thus the model has better chance to learn from cross-view information.
On the other hand, categories with super simple geometry (\eg speaker, cellphone) do not help to improve other categories, even not for themselves.
On the whole, MDN shows good generalization capability across semantic categories.

\subsubsection{Number of Views}
We then test how MDN performs \wrt the number of input views.
In Tab. \ref{tbl:view_number}, we see that MDN consistently performs better when more input views are available, even though the number of view is fixed as 3 for efficiency during the training.
This indicates that features from multiple views are well encoded in the statistics, and MDN is able to exploit additional information when seeing more images.
For reference, we train five MDNs with the input view number fixed at 2 to 5 respectively. 
As shown in Tab. \ref{tbl:view_number} ``Resp.'', the 3-view MDN performs very close to models trained with more views (\eg 4 and 5), which shows the model learns efficiently from fewer number of views during the training.
The 3-view MDN also outperform models trained with less views (\eg 2), which indicates additional information provided during the training can be effectively activated during the test even when observation is limited.
Overall, MDN generalizes well to different number of inputs.

\begin{table}[h]
    \begin{center}
    \caption{{\bf Performance \wrt Input View Numbers.} \textcolor{black}{`Resp.' means that MDN is trained with the same views as the testing. MDN performs consistently better when more view is given, even trained using only 3 views.}}
    \resizebox{0.85\columnwidth}{!}{
        \begin{tabular}{cccccc}
        \toprule
        \#train & \#test & 2 & 3 & 4 & 5 \\
        \midrule
        \midrule
        \multirow{3}{*}{3} & F-score($\tau$) $\uparrow$ & 64.48 & 66.44 & 67.66 & 68.29 \\
        & F-score($2 \tau$) $\uparrow$ & 78.74 & 80.33 & 81.36 & 81.97 \\
        & CD $\downarrow$ & 0.515 & 0.484 & 0.468 & 0.459\\
        \midrule
        \multirow{3}{*}{Resp.} & F-score($\tau$) $\uparrow$& 64.11 & 66.44 & 68.54 & 68.82 \\
        & F-score($2 \tau$) $\uparrow$ & 78.34 & 80.33 & 81.56 & 81.99 \\
        & CD $\downarrow$ & 0.527 & 0.484 & 0.467 & 0.452\\
        \bottomrule
        \end{tabular}
        }
    \end{center}
\label{tbl:view_number}
\end{table}

\subsection{Model Robustness Analysis}
In this section, we analyze the robustness of the model in many aspects. 
We analyze the model's ability to deal with initialization noise, cameras position error, and instance level optimization during test with differentiable renderer when silhouette mask are available.
Through these analyses, we explore why multiple views help for 3D reconstruction. 

\subsubsection{Initialization}
Firstly, we test if the model overfits to the input initialization, i.e. the MVP2M.
To this end, we add translation and random noise to the rough shape from MVP2M. 
We also take as input the mesh converted from 3DR2N2 using marching cubes algorithm \cite{LorensenC87}.
As shown in Fig. \ref{fig:init}, MDN successfully removes the noise, aligns the input with ground truth, and adds significant geometry details.
This experiments show that MDN is tolerant to input variance.


\subsubsection{Camera Pose}
We compare the performance of our method on camera pose estimation with Insafutdinov \etal~\cite{insafutdinov2018unsupervised} and Xu \etal~\cite{NIPS2019_DISN}. Given a point cloud in world space, we transform this point cloud using ground truth camera parameters and predicted parameters to camera view respectively. We first calculate the the mean distance in 3D space which is called $d_{3D}$. We then project the point cloud onto the input image plane using known intrinsics to compute the 2D reprojection error $d_{2D}$. We also evaluate 
Acc$_{2D}$ and ADD$_{3D}$ metrics.
With the help of continuous rotation and compact 1D translation representation, our method outperforms Insafutdinov \etal~\cite{insafutdinov2018unsupervised} and Xu \etal~\cite{NIPS2019_DISN} in terms of $d_{2D}$, Acc$_{2D}$ and ADD$_{3D}$. And our method achieves comparable results on $d_{3D}$.

\begin{table}[h]
    \begin{center}
    \caption{{\bf Quantitative Camera Pose Estimation Comparison.} We compare our method with Zhou \etal~\cite{zhou2019continuity} and DISN~\cite{NIPS2019_DISN}.}
    \resizebox{0.8\columnwidth}{!}{
    \begin{tabular}{ccccc}
        \toprule
        Methods  & $d_{2D}$ $\downarrow$ & $d_{3D}$ $\downarrow$ & Acc$_{2D}$ $\uparrow$ & ADD$_{3D}$ $\uparrow$\\
        \midrule
        \midrule
        Zhou \etal~\cite{zhou2019continuity} & 4.86 & 0.073 & - & -\\
        DISN~\cite{NIPS2019_DISN} & 2.95 & \textbf{0.047} & 73.34 & 87.31\\
        Ours & \textbf{2.39} & 0.049 & \textbf{74.32} & \textbf{93.32}\\
        \bottomrule
    \end{tabular}
    }
    \end{center}
    \label{tbl:campose}
\end{table}

Moreover, we test whether MDN is able to produce plausible 3D shape when camera pose has error. Since gathering image features require reasonable camera pose, using random camera pose is not feasible. Instead, we use camera pose estimation network to obtain reasonable camera extrinsics. 
As shown in Tab.~\ref{tbl:robustness} (w/ pred cam), even using predicted camera parameters, our MDN can also refine the initial mesh shape and improve performance compared to the baseline algorithms. 
Benefiting from more visual cues provided by multi-view images, the above experimental results show that MDN has a certain degree of robustness to camera pose errors.

\begin{table}
\begin{centering}
\caption{ {\bf Robustness of Different Methods.} pc, gtc, and s are short for predicted camera pose, ground-truth camera pose, and sihouette.  CD:Chamfer Distance. \label{tbl:robustness}}
\begin{tabu}to \columnwidth {X[1.5,l]X[2,c]X[1.5,c]X[1.5,c]X[1.5,c]X[2,c]X[1.5,c]}
\toprule 
\multirow{2}{*}{Methods} & \multirow{2}{*}{w/pc} & \multirow{2}{*}{w/gtc} & \multirow{2}{*}{w/s} & \multicolumn{2}{c}{F-score $\uparrow$} & \multirow{2}{*}{CD$\downarrow$}\tabularnewline
\cline{5-6}
 &  &  &  & $\tau$ & 2$\tau$ & \tabularnewline
\midrule
\midrule
\multirow{4}{*}{MVP2M} & $\surd$ &  &  & {\small{}59.93} & {\small{}75.85} & {\small{}0.482}\tabularnewline
 &  & $\surd$ &  & {\small{}61.20} & {\small{}76.96} & {\small{}0.456}\tabularnewline
 & $\surd$ &  & $\surd$ & {\small{}60.67} & {\small{}76.41} & {\small{}0.472}\tabularnewline
 &  & $\surd$ & $\surd$ & {\small{}62.80} & {\small{}78.01} & {\small{}0.437}\tabularnewline
\midrule 
\multirow{4}{*}{Ours-P} & $\surd$ &  &  & {\small{}63.22} & {\small{}78.52} & {\small{}0.428}\tabularnewline
 &  & $\surd$ &  & {\small{}67.32} & {\small{}81.22} & {\small{}0.381}\tabularnewline
 & $\surd$ &  & $\surd$ & {\small{}63.56} & {\small{}78.89} & {\small{}0.420}\tabularnewline
 &  & $\surd$ & $\surd$ & {\small{}68.45} & {\small{}82.19} & {\small{}0.360}\tabularnewline
\midrule 
\multirow{4}{*}{MVDISN} & $\surd$ &  &  & {\small{}67.86} & {\small{}80.09} & {\small{}0.553}\tabularnewline
 &  & $\surd$ &  & {\small{}70.78} & {\small{}82.22} & {\small{}0.439}\tabularnewline
 & $\surd$ &  & $\surd$ & {\small{}68.39} & {\small{}80.38} & {\small{}0.545}\tabularnewline
 &  & $\surd$ & $\surd$ & {\small{}71.78} & {\small{}82.71} & {\small{}0.429}\tabularnewline
\midrule 
\multirow{4}{*}{Ours-D} & $\surd$ &  &  & {\small{}70.07} & {\small{}81.79} & {\small{}0.518}\tabularnewline
 &  & $\surd$ &  & {\small{}75.24} & {\small{}85.04} & {\small{}0.390}\tabularnewline
 & $\surd$ &  & $\surd$ & {\small{}70.26} & {\small{}81.95} & {\small{}0.511}\tabularnewline
 &  & $\surd$ & $\surd$ & {\small{}75.74} & {\small{}85.33} & {\small{}0.383}\tabularnewline
\bottomrule 
\end{tabu}
\par\end{centering}
\end{table}

\subsubsection{Differentiable Renderer}
We test the effect of our method combined with the differentiable renderer. The differentiable renderer can perform test-time optimization for 3D shape during inference.


In this section, we use the differentiable renderer during inference. The 3D shapes can be further optimized by matching the silhouette of the predicted shapes with the multi-view input images.
We use 2D silhouette loss for the optimization, so that the generated 3D model has a better alignment at the views corresponding to the input images.
Optimization at the image level helps to restore the unique geometric details of the 3D shape. Moreover, the optimization process can make the shape and the input image align better.
Since the previous method based on the differentiable renderer either requires a large number of multi-view silhouettes as supervision~\cite{liu2020general}, or processes a single color image~\cite{chen2019dibr}, it is not suitable for our task.
On the basis of Liu \etal~\cite{liu2020general}, we only use the part of its differentiable renderer. We use three input images with corresponding silhouettes and network output 3D mesh models for optimization.

\begin{figure}[ht]
	\centering
	\includegraphics[width=0.95\columnwidth]{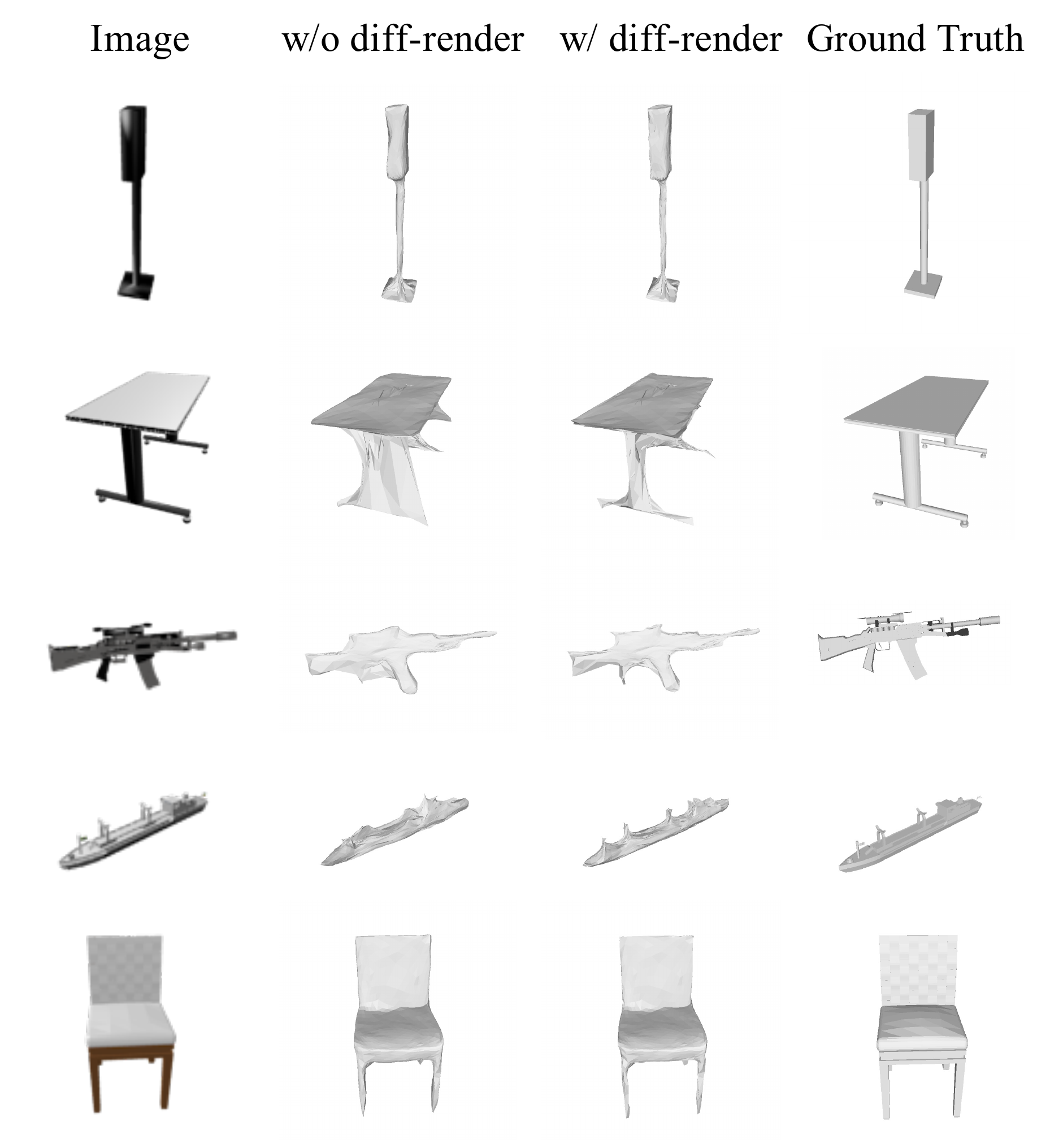}
	\caption{{\bf Qualitative Examples with or without differentiable renderer.}}
	\label{fig:diffren}
\end{figure}

As shown in Tab.~\ref{tbl:robustness} (w/ silhouette), through the optimization leveraging the differentiable renderer, the performance of our method can be further improved. The qualitative results are shown in the Fig.~\ref{fig:diffren}. It can be seen that using the differentiable renderer can further optimize the predicted 3D shape, especially geometric details such as larger corners.

\begin{figure}[ht]
	\centering
	\includegraphics[width=0.9\columnwidth]{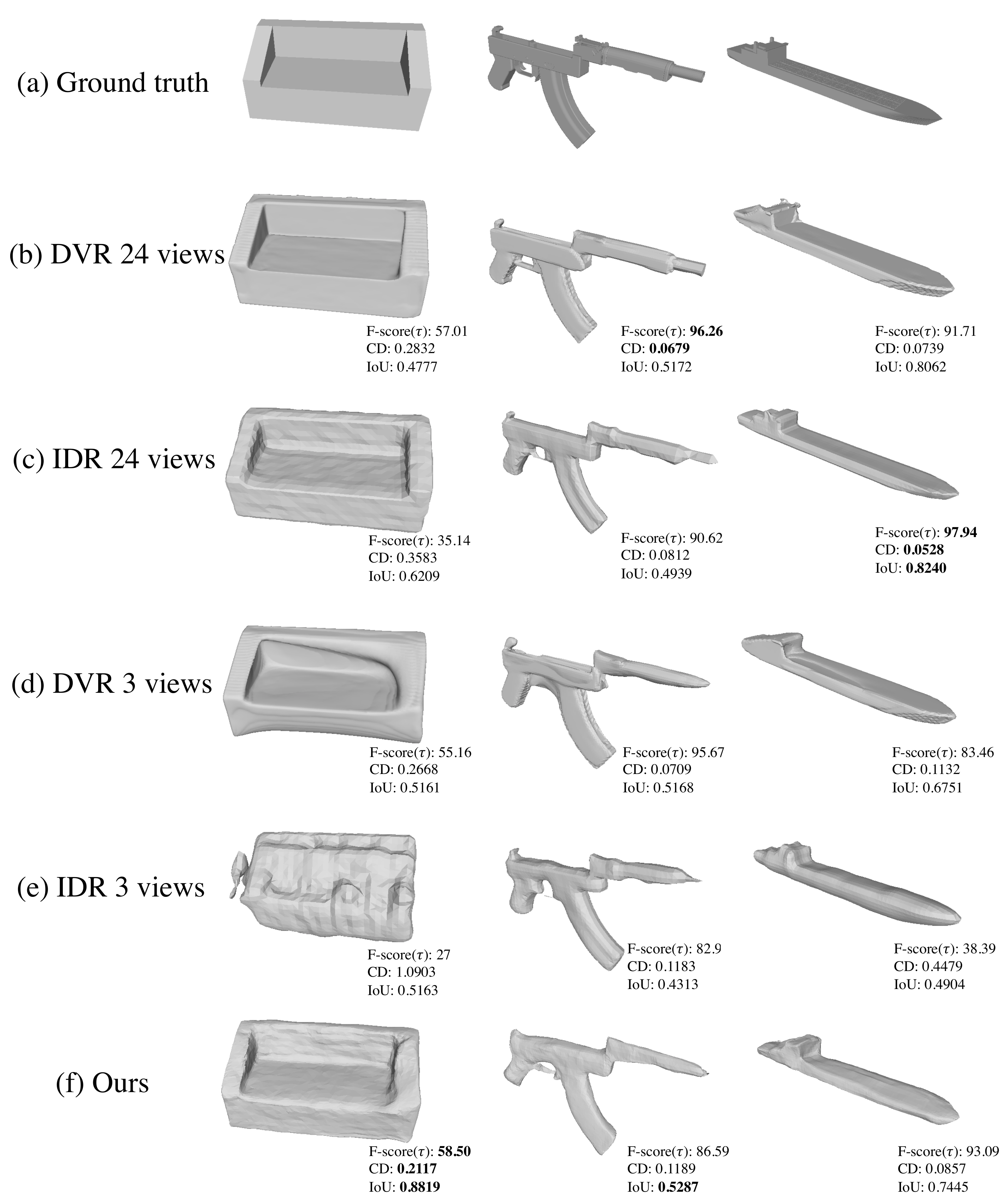}
	\caption{\textcolor{black}{{\bf Qualitative and Quantitative Results Compared with DVR~\cite{niemeyer2020differentiable} and IDR~\cite{yariv2020multiview}.} Note that our method only takes 3-view inputs. DVR and IDR are evaluated with the same or more input views. 
	}}
	\label{fig:render_comparison}
\end{figure}



\textcolor{black}{
\subsubsection{Comparison of Neural Rendering Approaches}
We also compare our method with representative neural rendering methods, i.e., Differentiable Volumetric Rendering (DVR)~\cite{niemeyer2020differentiable} and Implicit Differentiable Rendering (IDR)~\cite{yariv2020multiview} in Fig.~\ref{fig:render_comparison} qualitatively. \textcolor{black}{Quantitative results of them are also shown in Fig.~\ref{fig:render_comparison}. 
Both DVR and IDR are trained with the instance-level, i.e., a separate model needs to be trained for each individual instance. Such online optimization is powerful in reconstruction, but suffers from poor efficiency to test a new coming instance. Because the computation with expensive inference time optimization is required.
We train these methods according to the official multi-view reconstruction setting.}
The initial shape of our method is MVDISN. Since our method only takes 3 images for each inference, we also compare the results of DVR and IDR with 3-view input images. Although rendering-based methods can achieve good results in Fig.~\ref{fig:render_comparison}(b)(c) with 24-view inputs, it costs several hours to train an instance-level neural rendering model for each sample, which is inefficient. Besides, if we reduce the input views to three, the performances of DVR and IDR are degraded obviously as shown in Fig~\ref{fig:render_comparison}(d)(e). And our method can get comparable results with 3-view input images more efficiently.
}

\begin{figure}[ht]
	\centering
	\includegraphics[width=0.9\columnwidth]{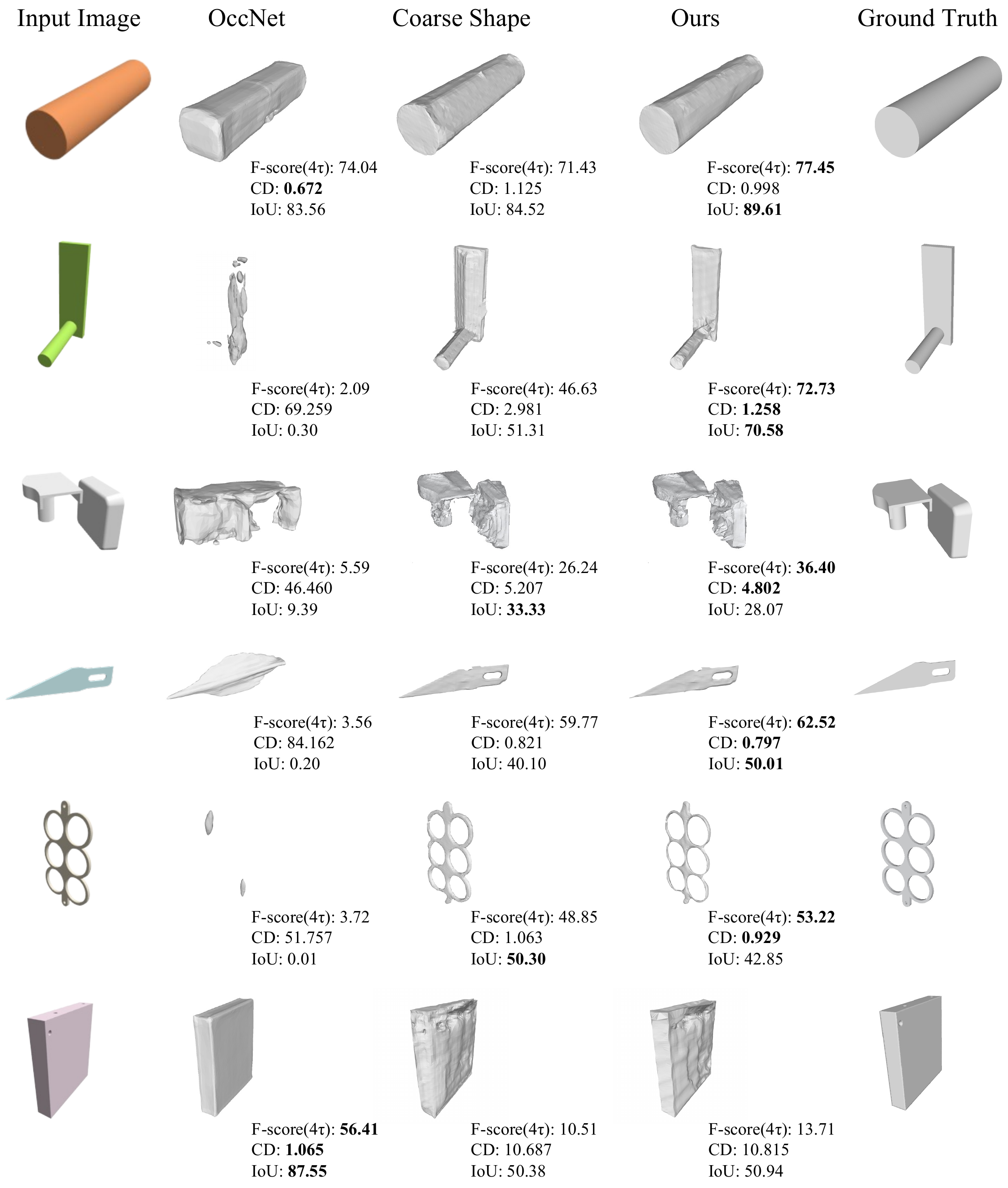}
	\caption{\textcolor{black}{{\bf Qualitative and Quantitative Results on ABC dataset~\cite{koch2019abc}.} We show the cross-dataset generalization ability. From left to right, input images, results from OccNet, coarse results from MVDISN, results from Our-D, and the ground truth. Our MDN can still improve results from the cross-dataset.}}
	\label{fig:abc}
\end{figure}

\textcolor{black}{
\subsubsection{Cross-dataset Generalization}
We evaluated the cross-data generalization ability of our method. \textcolor{black}{We train models on ShapeNet and test them on the ABC dataset~\cite{koch2019abc} directly without any finetuning for the generalization analysis. Qualitative and quantitative results are shown in Fig.~\ref{fig:abc}.} 
Thanks to feature learning from multiple views, our coarse shape generation from MVDISN has certain cross-data generalization capabilities. \textcolor{black}{And our MDN can further refine the 3D shape in both qualitative and quantitative results, especially when the coarse shape has geometry noise. Besides, the single-view based OccNet~\cite{mescheder2019occupancy} may fails to achieve reliable results in ABC dataset.}
Nevertheless, generalization across datasets is a very challenging task. When the geometric shape of the new dataset is complex or the coarse shape generation stage failed, the performance of our approach will degrade (\eg the third example in Fig.~\ref{fig:abc}). }

\subsection{Ablation Study}
In this section, we verify the qualitative and quantitative improvements from statistic feature pooling, re-sampled Chamfer distance, and iterative refinement.

\subsubsection{Statistical Feature}\label{sec:stat_feat}
We first check the importance of using feature statistics. We train MDN with the ordinary concatenation. This maintains all the features loss-less to potentially produce better geometry, but does not support variable number of inputs any more.
Surprisingly, our model with feature statistics (Tab. \ref{tbl:ablation}, ``Full Model'') still outperforms the one with concatenation (Tab. \ref{tbl:ablation}, ``-Feat Stat'').
This is probably because that our feature statistics is invariant to the input order, such that the network learns more efficiently during the training. It also explicitly encodes cross-view feature correlations, which can be directly leveraged by the network.


\begin{figure}[ht]
	\centering
	\includegraphics[width=\columnwidth]{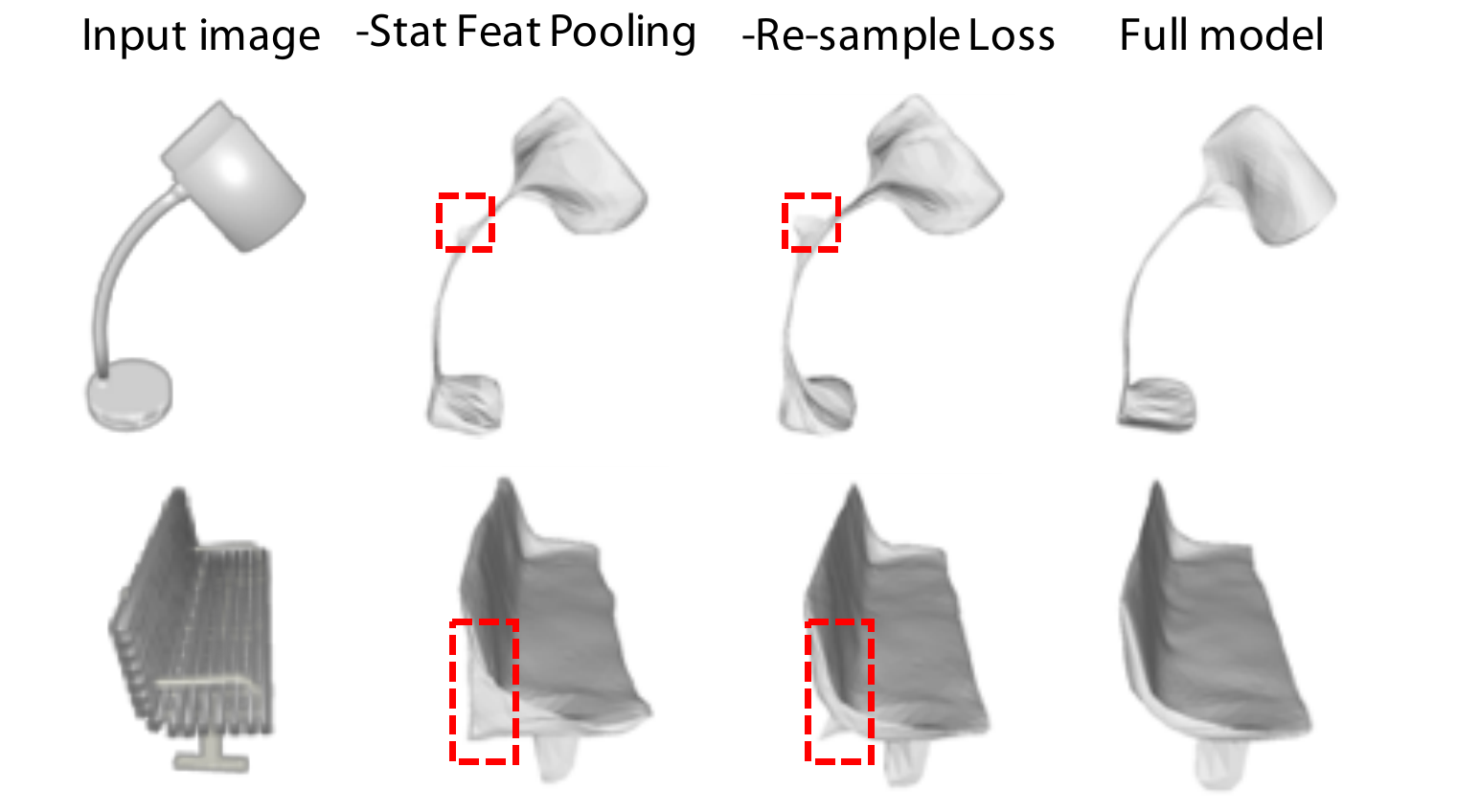}
	\caption{{\bf Qualitative Ablation Study.} We show meshes from the MDN with statistics feature or re-sampling loss disabled.}
	\label{Fig.ablation}
\end{figure}

\begin{table}[h]
    \begin{center}
    \caption{{\bf Quantitative Ablation Study.} We show the metrics of the MDN with statistics feature or re-sampling loss disabled.}
    \resizebox{0.85\columnwidth}{!}{
    \begin{tabular}{cccc}
        \toprule
        Metrics & F-score($\tau$) $\uparrow$ & F-score($2\tau$) $\uparrow$ & CD $\downarrow$\\
        \midrule
        \midrule
        -Feat Stat & 65.26 & 79.13 & 0.511\\
        -Re-sample Loss & 66.26 & 80.04 & 0.496\\
        Full Model & \textbf{67.32} & \textbf{81.22} & \textbf{0.381} \\
        \bottomrule
    \end{tabular}
    }
    \end{center}
    \label{tbl:ablation}
\end{table}

\subsubsection{Re-sampled Chamfer Distance}
We then investigate the impact of the re-sampled Chamfer loss.
We train our model using the traditional Chamfer loss only on mesh vertices as defined in Pixel2Mesh, and all metrics drop consistently (Tab. \ref{tbl:ablation}, ``-Re-sample Loss''). 
Intuitively, our re-sampling loss is especially helpful for places with sparse vertices and irregular faces, such as the elongated lamp neck as shown in Fig. \ref{Fig.ablation}, 3rd column.
It also prevents big mistakes from happening on a single vertex, \eg the spike on bench, where our loss penalizes a lot of sampled points on wrong faces caused by the vertex but the standard Chamfer loss only penalizes one point.



\begin{figure}[t]
	\centering
	\includegraphics[width=\columnwidth]{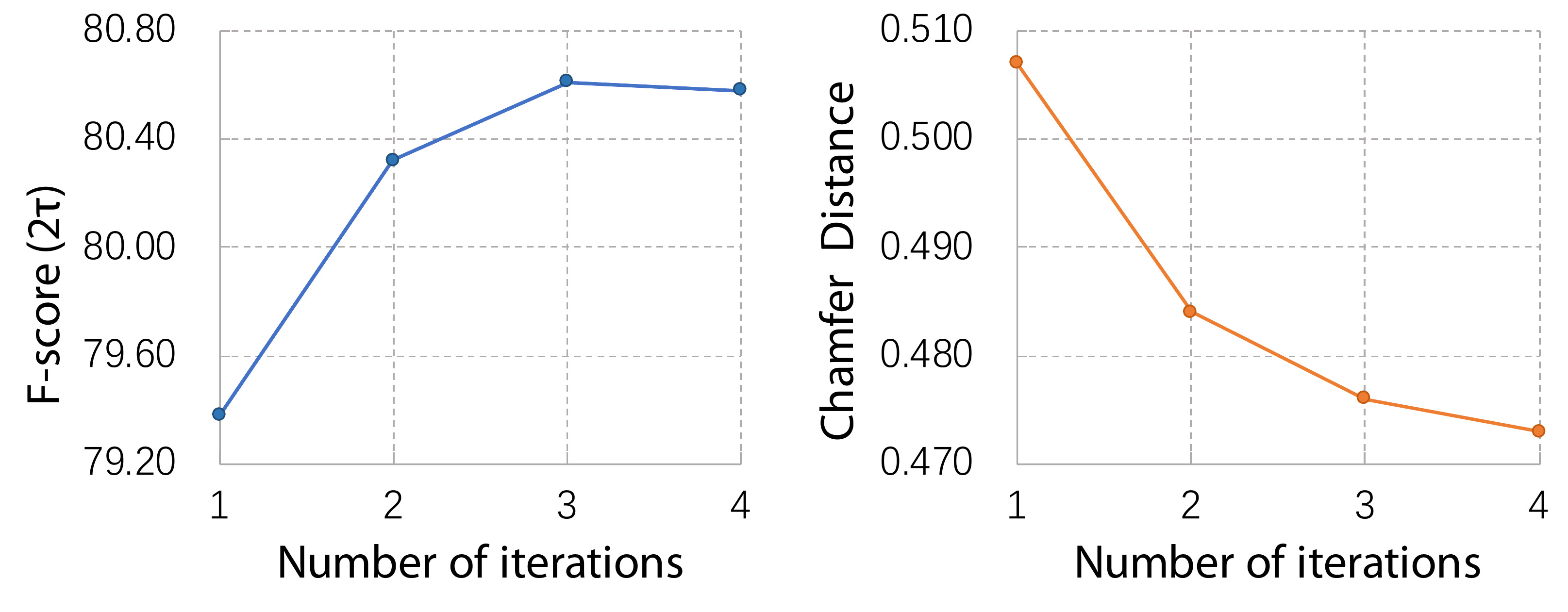}
	\caption{{\bf Performance with Different Iterations. } The performance keeps improving with more iterations and saturate at three.
    }
	\label{fig:iter}
\end{figure}

\subsubsection{Number of Iteration}
Figure \ref{fig:iter} shows that the performance of our model keeps improving with more iterations, and is roughly saturated at three.
Therefore we choose to run three iterations during the inference even though marginal improvements can be further obtained from more iterations.

\begin{figure}[t]
	\centering
	\includegraphics[width=\columnwidth]{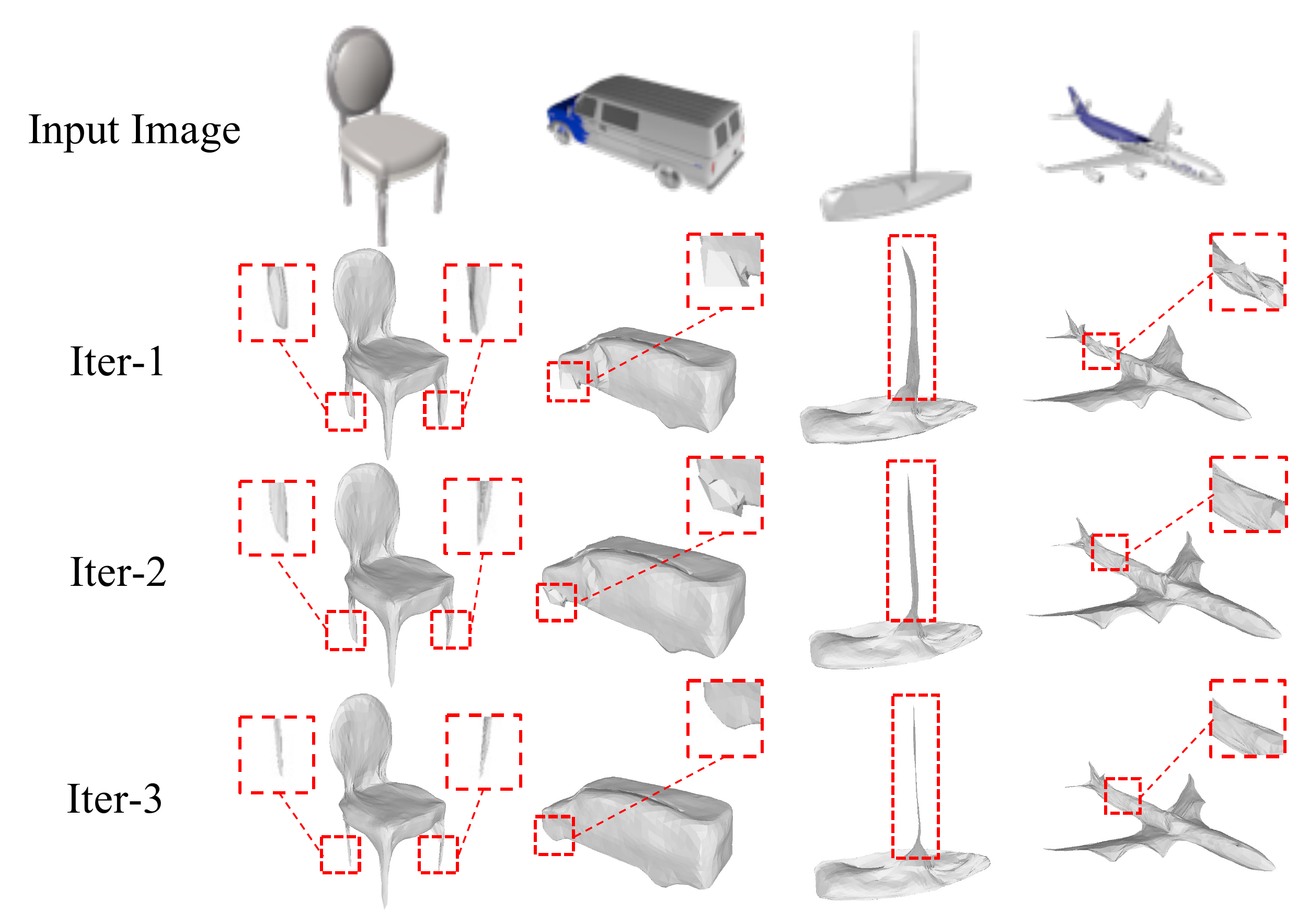}
	\caption{{\bf Qualitative Examples with Different Iterations.} }
	\label{fig.iter2}
\end{figure}

\begin{figure}[ht]
	\centering
	\includegraphics[width=\columnwidth]{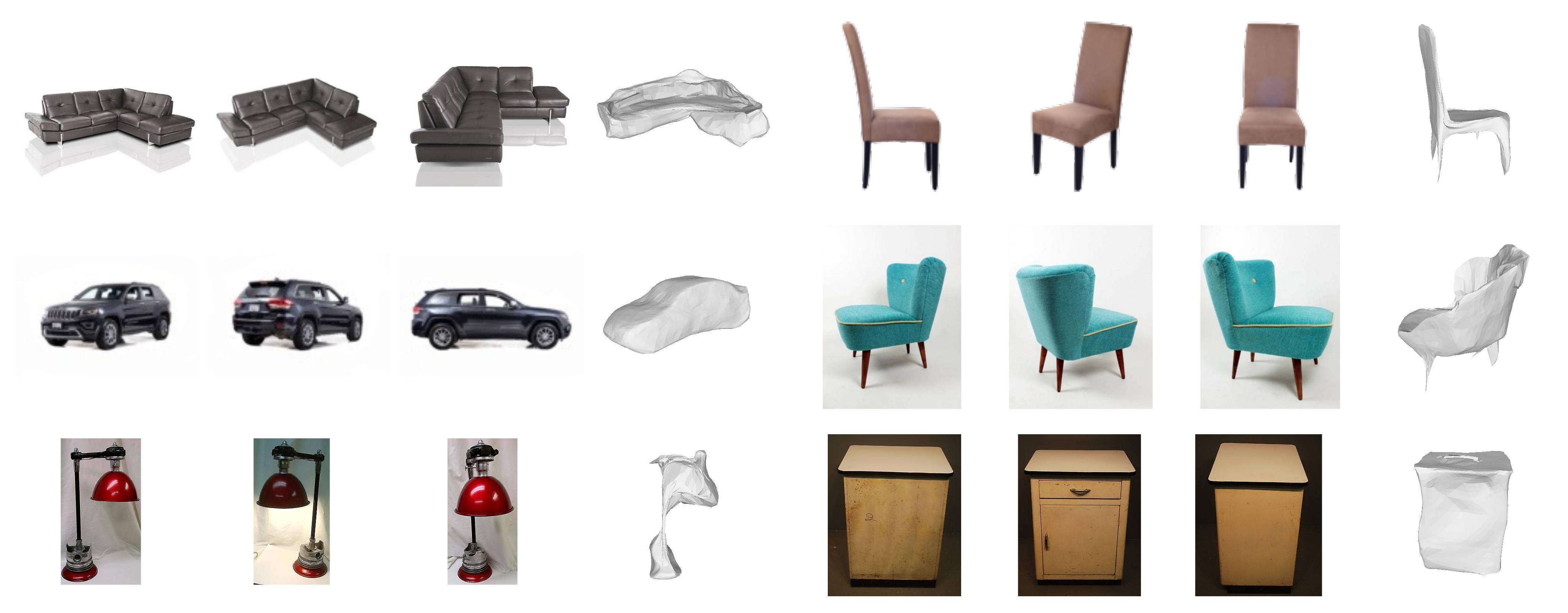}
	\caption{{\bf Qualitative results of real-world images from the online products dataset and Internet.}}
	\label{fig:real}
\end{figure}

\begin{figure}[ht]
	\centering
	\includegraphics[width=0.85\columnwidth]{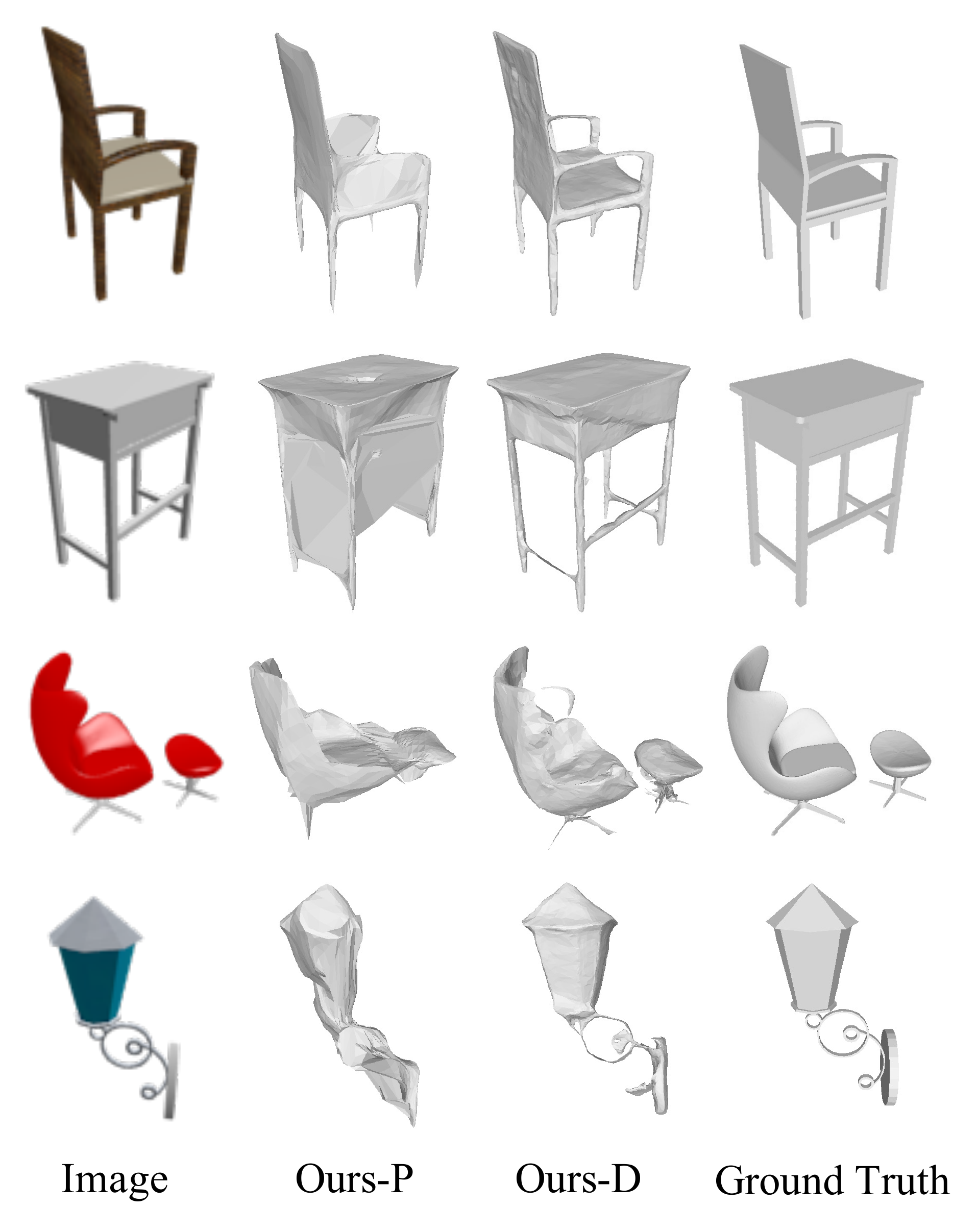}
	\caption{{\bf Comparison of different coarse initial shape method.}}
	\label{fig:topo}
\end{figure}

\section{Conclusion}
We propose a graph convolutional framework to produce 3D mesh model from multiple images.
Our model learns to exploit cross-view information and generates vertex deformation iteratively to improve the mesh produced from the direct prediction methods, \eg Pixel2Mesh and its multi-view extension.
Inspired by multi-view geometry methods, our model searches in the nearby area around each vertex for an optimal place to relocate it.
Compared to previous works, our model achieves the state-of-the-art performance, produces shapes containing accurate surface details rather than merely visually plausible from input views, and shows good generalization capability in many aspects.
For future work, combining with efficient shape retrieval for initialization, integrating with multi-view stereo models for explicit photometric consistency, and extending to scene scales are some of the practical directions to explore.
On a high level, how to integrating the similar idea in emerging new representations, such as part based model with shape basis and learned function \cite{Park_2019_CVPR} are interesting for further study.

\ifCLASSOPTIONcompsoc
  \section*{Acknowledgments}
\else
  \section*{Acknowledgment}
\fi

This work was supported in part by the National Natural Science Foundation of China Grant (62076067),  SMSTM Project (2021SHZDZX0103),  Shanghai Municipal Science and Technology Major Project (No.2018SHZDZX01), ZJ Lab, and Shanghai Center for Brain Science and Brain-Inspired Technology. Dr. Yanwei Fu is also with and Shanghai Key Lab of Intelligent Information Processing, and Fudan ISTBI—ZJNU Algorithm Centre for Brain-inspired Intelligence, Zhejiang Normal University, Jinhua, China.

\ifCLASSOPTIONcaptionsoff
  \newpage
\fi




\bibliographystyle{IEEEtran}
\bibliography{egbib}

\begin{thebibliography}{10}
\providecommand{\url}[1]{#1}
\csname url@samestyle\endcsname
\providecommand{\newblock}{\relax}
\providecommand{\bibinfo}[2]{#2}
\providecommand{\BIBentrySTDinterwordspacing}{\spaceskip=0pt\relax}
\providecommand{\BIBentryALTinterwordstretchfactor}{4}
\providecommand{\BIBentryALTinterwordspacing}{\spaceskip=\fontdimen2\font plus
\BIBentryALTinterwordstretchfactor\fontdimen3\font minus
  \fontdimen4\font\relax}
\providecommand{\BIBforeignlanguage}[2]{{%
\expandafter\ifx\csname l@#1\endcsname\relax
\typeout{** WARNING: IEEEtran.bst: No hyphenation pattern has been}%
\typeout{** loaded for the language `#1'. Using the pattern for}%
\typeout{** the default language instead.}%
\else
\language=\csname l@#1\endcsname
\fi
#2}}
\providecommand{\BIBdecl}{\relax}
\BIBdecl

\bibitem{wang2018pixel2mesh}
N.~Wang, Y.~Zhang, Z.~Li, Y.~Fu, W.~Liu, and Y.-G. Jiang, ``Pixel2mesh:
  Generating 3d mesh models from single rgb images,'' in \emph{Proceedings of
  the European Conference on Computer Vision (ECCV)}, 2018.

\bibitem{HarlteyZ2001}
A.~Harltey and A.~Zisserman, \emph{Multiple view geometry in computer vision
  {(2.} ed.)}.\hskip 1em plus 0.5em minus 0.4em\relax Cambridge University
  Press, 2006.

\bibitem{KendallMDH17}
A.~Kendall, H.~Martirosyan, S.~Dasgupta, and P.~Henry, ``End-to-end learning of
  geometry and context for deep stereo regression,'' in \emph{Proceedings of
  the IEEE International Conference on Computer Vision (ICCV)}, 2017, pp.
  66--75.

\bibitem{NIPS2019_DISN}
Q.~Xu, W.~Wang, D.~Ceylan, R.~Mech, and U.~Neumann, ``Disn: Deep implicit
  surface network for high-quality single-view 3d reconstruction,'' in
  \emph{Advances in Neural Information Processing Systems (NeurIPS)}, 2019, pp.
  492--502.

\bibitem{wen2019pixel2mesh++}
C.~Wen, Y.~Zhang, Z.~Li, and Y.~Fu, ``Pixel2mesh++: Multi-view 3d mesh
  generation via deformation,'' in \emph{Proceedings of the IEEE International
  Conference on Computer Vision (ICCV)}, 2019, pp. 1042--1051.

\bibitem{choy20163d}
C.~B. Choy, D.~Xu, J.~Gwak, K.~Chen, and S.~Savarese, ``3d-r2n2: A unified
  approach for single and multi-view 3d object reconstruction,'' in
  \emph{Proceedings of the European Conference on Computer Vision (ECCV)},
  2016.

\bibitem{wang2017cnn}
P.-S. Wang, Y.~Liu, Y.-X. Guo, C.-Y. Sun, and X.~Tong, ``O-cnn: Octree-based
  convolutional neural networks for 3d shape analysis,'' \emph{ACM Transactions
  on Graphics (TOG)}, vol.~36, no.~4, p.~72, 2017.

\bibitem{qi2016pointnet}
C.~R. Qi, H.~Su, K.~Mo, and L.~J. Guibas, ``Pointnet: Deep learning on point
  sets for 3d classification and segmentation,'' in \emph{Proceedings of the
  IEEE Conference on Computer Vision and Pattern Recognition (CVPR)}, 2017.

\bibitem{fan2017point}
H.~Fan, H.~Su, and L.~J. Guibas, ``A point set generation network for 3d object
  reconstruction from a single image,'' in \emph{Proceedings of the IEEE
  Conference on Computer Vision and Pattern Recognition (CVPR)}, 2017.

\bibitem{QiLWSG18}
C.~R. Qi, W.~Liu, C.~Wu, H.~Su, and L.~J. Guibas, ``Frustum pointnets for 3d
  object detection from {RGB-D} data,'' in \emph{Proceedings of the IEEE
  Conference on Computer Vision and Pattern Recognition (CVPR)}, 2018.

\bibitem{KatoUH18}
H.~Kato, Y.~Ushiku, and T.~Harada, ``Neural 3d mesh renderer,'' in
  \emph{Proceedings of the IEEE Conference on Computer Vision and Pattern
  Recognition (CVPR)}, 2018.

\bibitem{SinhaUHR17}
A.~Sinha, A.~Unmesh, Q.~Huang, and K.~Ramani, ``Surfnet: Generating 3d shape
  surfaces using deep residual networks,'' in \emph{Proceedings of the IEEE
  Conference on Computer Vision and Pattern Recognition (CVPR)}, 2017.

\bibitem{TatarchenkoDB16}
M.~Tatarchenko, A.~Dosovitskiy, and T.~Brox, ``Multi-view 3d models from single
  images with a convolutional network,'' in \emph{Proceedings of the European
  Conference on Computer Vision (ECCV)}, 2016.

\bibitem{Richter018}
S.~R. Richter and S.~Roth, ``Matryoshka networks: Predicting 3d geometry via
  nested shape layers,'' in \emph{Proceedings of the IEEE Conference on
  Computer Vision and Pattern Recognition (CVPR)}, 2018.

\bibitem{mescheder2019occupancy}
L.~Mescheder, M.~Oechsle, M.~Niemeyer, S.~Nowozin, and A.~Geiger, ``Occupancy
  networks: Learning 3d reconstruction in function space,'' in
  \emph{Proceedings of the IEEE Conference on Computer Vision and Pattern
  Recognition (CVPR)}, 2019, pp. 4460--4470.

\bibitem{chen2019learning}
Z.~Chen and H.~Zhang, ``Learning implicit fields for generative shape
  modeling,'' in \emph{Proceedings of the IEEE Conference on Computer Vision
  and Pattern Recognition (CVPR)}, 2019, pp. 5939--5948.

\bibitem{Liao2018CVPR}
Y.~Liao, S.~Donne, and A.~Geiger, ``Deep marching cubes: Learning explicit
  surface representations,'' in \emph{Proceedings of the IEEE Conference on
  Computer Vision and Pattern Recognition (CVPR)}, 2018, pp. 2916--1925.

\bibitem{atzmon2019controlling}
M.~Atzmon, N.~Haim, L.~Yariv, O.~Israelov, H.~Maron, and Y.~Lipman,
  ``Controlling neural level sets,'' in \emph{Advances in Neural Information
  Processing Systems (NeurIPS)}, 2019, pp. 2032--2041.

\bibitem{SAL_Atzmon_2020_CVPR}
M.~Atzmon and Y.~Lipman, ``Sal: Sign agnostic learning of shapes from raw
  data,'' in \emph{Proceedings of the IEEE Conference on Computer Vision and
  Pattern Recognition (CVPR)}, 2020, pp. 2565--2574.

\bibitem{Park_2019_CVPR}
J.~J. Park, P.~Florence, J.~Straub, R.~Newcombe, and S.~Lovegrove, ``Deepsdf:
  Learning continuous signed distance functions for shape representation,'' in
  \emph{Proceedings of the IEEE Conference on Computer Vision and Pattern
  Recognition (CVPR)}, June 2019.

\bibitem{DurouFS08}
J.~Durou, M.~Falcone, and M.~Sagona, ``Numerical methods for
  shape-from-shading: {A} new survey with benchmarks,'' \emph{Computer Vision
  and Image Understanding}, vol. 109, no.~1, pp. 22--43, 2008.

\bibitem{ZhangTCS99}
R.~Zhang, P.~Tsai, J.~E. Cryer, and M.~Shah, ``Shape from shading: {A}
  survey,'' \emph{{IEEE} Trans. Pattern Anal. Mach. Intell.}, vol.~21, no.~8,
  pp. 690--706, 1999.

\bibitem{MarinosB90}
C.~Marinos and A.~Blake, ``Shape from texture: the homogeneity hypothesis,'' in
  \emph{Proceedings of the IEEE International Conference on Computer Vision
  (ICCV)}, 1990, pp. 350--353.

\bibitem{FavaroS05}
P.~Favaro and S.~Soatto, ``A geometric approach to shape from defocus,''
  \emph{{IEEE} Trans. Pattern Anal. Mach. Intell.}, vol.~27, no.~3, pp.
  406--417, 2005.

\bibitem{GirdharFRG16}
R.~Girdhar, D.~F. Fouhey, M.~Rodriguez, and A.~Gupta, ``Learning a predictable
  and generative vector representation for objects,'' in \emph{Proceedings of
  the European Conference on Computer Vision (ECCV)}, 2016.

\bibitem{WuZXFT16}
J.~Wu, C.~Zhang, T.~Xue, B.~Freeman, and J.~Tenenbaum, ``Learning a
  probabilistic latent space of object shapes via 3d generative-adversarial
  modeling,'' in \emph{Advances in Neural Information Processing Systems
  (Neurips)}, 2016.

\bibitem{HaneTM17}
C.~Hane, S.~Tulsiani, and J.~Malik, ``Hierarchical surface prediction for 3d
  object reconstruction,'' in \emph{Proceedings of the International Conference
  on 3D Vision (3DV)}, 2017.

\bibitem{RieglerUG17}
G.~Riegler, A.~O. Ulusoy, and A.~Geiger, ``Octnet: Learning deep 3d
  representations at high resolutions,'' in \emph{Proceedings of the IEEE
  Conference on Computer Vision and Pattern Recognition (CVPR)}, 2017.

\bibitem{TatarchenkoDB17}
M.~Tatarchenko, A.~Dosovitskiy, and T.~Brox, ``Octree generating networks:
  Efficient convolutional architectures for high-resolution 3d outputs,'' in
  \emph{Proceedings of the IEEE International Conference on Computer Vision
  (ICCV)}, 2017.

\bibitem{JohnstonGCR17}
A.~Johnston, R.~Garg, G.~Carneiro, and I.~D. Reid, ``Scaling cnns for high
  resolution volumetric reconstruction from a single image,'' in
  \emph{Proceedings of the IEEE International Conference on Computer Vision
  (ICCV)}, 2017.

\bibitem{groueix2018AtlasNet}
T.~Groueix, M.~Fisher, V.~G. Kim, B.~Russell, and M.~Aubry, ``{AtlasNet: A
  Papier-M\^ach\'e Approach to Learning 3D Surface Generation},'' in
  \emph{Proceedings of the IEEE Conference on Computer Vision and Pattern
  Recognition (CVPR)}, 2018, pp. 216--224.

\bibitem{goel2020shape}
S.~Goel, A.~Kanazawa, and J.~Malik, ``Shape and viewpoint without keypoints,''
  in \emph{Proceedings of the European Conference on Computer Vision (ECCV)},
  2020, pp. 88--104.

\bibitem{ye2021shelf}
Y.~Ye, S.~Tulsiani, and A.~Gupta, ``Shelf-supervised mesh prediction in the
  wild,'' in \emph{Proceedings of the IEEE Conference on Computer Vision and
  Pattern Recognition (CVPR)}, 2021, pp. 8843--8852.

\bibitem{gupta2020neuralmeshflow}
K.~Gupta and M.~Chandraker, ``Neural mesh flow: 3d manifold mesh generation via
  diffeomorphic flows,'' in \emph{Advances in Neural Information Processing
  Systems (NeurIPS)}, 2020, pp. 1747--1758.

\bibitem{TulsianiSGEM17}
S.~Tulsiani, H.~Su, L.~J. Guibas, A.~A. Efros, and J.~Malik, ``Learning shape
  abstractions by assembling volumetric primitives,'' in \emph{Proceedings of
  the IEEE Conference on Computer Vision and Pattern Recognition (CVPR)}, 2017.

\bibitem{Niu0018}
C.~Niu, J.~Li, and K.~Xu, ``Im2struct: Recovering 3d shape structure from a
  single {RGB} image,'' in \emph{Proceedings of the IEEE Conference on Computer
  Vision and Pattern Recognition (CVPR)}, 2018.

\bibitem{paschalidou2019superquadrics}
D.~Paschalidou, A.~O. Ulusoy, and A.~Geiger, ``Superquadrics revisited:
  Learning 3d shape parsing beyond cuboids,'' in \emph{Proceedings of the IEEE
  Conference on Computer Vision and Pattern Recognition (CVPR)}, 2019, pp.
  10\,344--10\,353.

\bibitem{genova2019learning}
K.~Genova, F.~Cole, D.~Vlasic, A.~Sarna, W.~T. Freeman, and T.~Funkhouser,
  ``Learning shape templates with structured implicit functions,'' in
  \emph{Proceedings of the IEEE/CVF International Conference on Computer
  Vision}, 2019, pp. 7154--7164.

\bibitem{deng2020cvxnet}
B.~Deng, K.~Genova, S.~Yazdani, S.~Bouaziz, G.~Hinton, and A.~Tagliasacchi,
  ``Cvxnet: Learnable convex decomposition,'' in \emph{Proceedings of the IEEE
  Conference on Computer Vision and Pattern Recognition (CVPR)}, 2020, pp.
  31--44.

\bibitem{paschalidou2021neural}
D.~Paschalidou, A.~Katharopoulos, A.~Geiger, and S.~Fidler, ``Neural parts:
  Learning expressive 3d shape abstractions with invertible neural networks,''
  in \emph{Proceedings of the IEEE Conference on Computer Vision and Pattern
  Recognition (CVPR)}, 2021, pp. 3204--3215.

\bibitem{TulsianiKCM17}
S.~Tulsiani, A.~Kar, J.~Carreira, and J.~Malik, ``Learning category-specific
  deformable 3d models for object reconstruction,'' \emph{{IEEE} Trans. Pattern
  Anal. Mach. Intell.}, vol.~39, no.~4, pp. 719--731, 2017.

\bibitem{KanazawaTEM18}
A.~Kanazawa, S.~Tulsiani, A.~A. Efros, and J.~Malik, ``Learning
  category-specific mesh reconstruction from image collections,'' in
  \emph{Proceedings of the European Conference on Computer Vision (ECCV)},
  2018.

\bibitem{HuangWK15}
Q.~Huang, H.~Wang, and V.~Koltun, ``Single-view reconstruction via joint
  analysis of image and shape collections,'' \emph{{ACM} Trans. Graph.},
  vol.~34, no.~4, pp. 87:1--87:10, 2015.

\bibitem{SuHMKG14}
H.~Su, Q.~Huang, N.~J. Mitra, Y.~Li, and L.~J. Guibas, ``Estimating image depth
  using shape collections,'' \emph{{ACM} Trans. Graph.}, vol.~33, no.~4, pp.
  37:1--37:11, 2014.

\bibitem{kurenkov2018deformnet}
A.~Kurenkov, J.~Ji, A.~Garg, V.~Mehta, J.~Gwak, C.~Choy, and S.~Savarese,
  ``Deformnet: Free-form deformation network for 3d shape reconstruction from a
  single image,'' in \emph{WACV}, 2018, pp. 858--866.

\bibitem{TatarchenkoRRLKB19}
M.~Tatarchenko, S.~Richter, R.~Ranftl, Z.~Li, V.~Koltun, and T.~Brox, ``What do
  single-view 3d reconstruction networks learn?'' in \emph{Proceedings of the
  IEEE Conference on Computer Vision and Pattern Recognition (CVPR)}, 2019.

\bibitem{yao2018mvsnet}
Y.~Yao, Z.~Luo, S.~Li, T.~Fang, and L.~Quan, ``Mvsnet: Depth inference for
  unstructured multi-view stereo,'' in \emph{Proceedings of the European
  Conference on Computer Vision (ECCV)}, 2018.

\bibitem{huang2018deepmvs}
P.-H. Huang, K.~Matzen, J.~Kopf, N.~Ahuja, and J.-B. Huang, ``Deepmvs: Learning
  multi-view stereopsis,'' in \emph{Proceedings of the IEEE Conference on
  Computer Vision and Pattern Recognition (CVPR)}, 2018.

\bibitem{im2018dpsnet}
S.~Im, H.-G. Jeon, S.~Lin, and I.~S. Kweon, ``Dpsnet: End-to-end deep plane
  sweep stereo,'' in \emph{International Conference on Learning Representations
  (ICLR)}, 2018.

\bibitem{zhang2018activestereonet}
Y.~Zhang, S.~Khamis, C.~Rhemann, J.~Valentin, A.~Kowdle, V.~Tankovich,
  M.~Schoenberg, S.~Izadi, T.~Funkhouser, and S.~Fanello, ``Activestereonet:
  End-to-end self-supervised learning for active stereo systems,'' in
  \emph{Proceedings of the European Conference on Computer Vision (ECCV)},
  2018, pp. 784--801.

\bibitem{donne2019learning}
S.~Donne and A.~Geiger, ``Learning non-volumetric depth fusion using successive
  reprojections,'' in \emph{Proceedings of the IEEE Conference on Computer
  Vision and Pattern Recognition (CVPR)}, 2019, pp. 7634--7643.

\bibitem{kar2017learning}
A.~Kar, C.~H{\"a}ne, and J.~Malik, ``Learning a multi-view stereo machine,'' in
  \emph{Advances in Neural Information Processing Systems (Neurips)}, 2017, pp.
  365--376.

\bibitem{GwakCCGS17}
J.~Gwak, C.~B. Choy, M.~Chandraker, A.~Garg, and S.~Savarese, ``Weakly
  supervised 3d reconstruction with adversarial constraint,'' in
  \emph{Proceedings of the International Conference on 3D Vision (3DV)}, 2017.

\bibitem{loper2014opendr}
M.~M. Loper and M.~J. Black, ``Opendr: An approximate differentiable
  renderer,'' in \emph{European Conference on Computer Vision}.\hskip 1em plus
  0.5em minus 0.4em\relax Springer, 2014, pp. 154--169.

\bibitem{xie2021neural}
Y.~Xie, T.~Takikawa, S.~Saito, O.~Litany, S.~Yan, N.~Khan, F.~Tombari,
  J.~Tompkin, V.~Sitzmann, and S.~Sridhar, ``Neural fields in visual computing
  and beyond,'' \emph{arXiv preprint arXiv: Arxiv-2111.11426}, 2021.

\bibitem{mildenhall2020nerf}
B.~Mildenhall, P.~P. Srinivasan, M.~Tancik, J.~T. Barron, R.~Ramamoorthi, and
  R.~Ng, ``Nerf: Representing scenes as neural radiance fields for view
  synthesis,'' in \emph{Proceedings of the European Conference on Computer
  Vision (ECCV)}, 2020, pp. 405--421.

\bibitem{niemeyer2020differentiable}
M.~Niemeyer, L.~Mescheder, M.~Oechsle, and A.~Geiger, ``Differentiable
  volumetric rendering: Learning implicit 3d representations without 3d
  supervision,'' in \emph{Proceedings of the IEEE Conference on Computer Vision
  and Pattern Recognition (CVPR)}, 2020, pp. 3504--3515.

\bibitem{yariv2020multiview}
L.~Yariv, Y.~Kasten, D.~Moran, M.~Galun, M.~Atzmon, B.~Ronen, and Y.~Lipman,
  ``Multiview neural surface reconstruction by disentangling geometry and
  appearance,'' vol.~33, pp. 2492--2502, 2020.

\bibitem{liu2020general}
S.~Liu, T.~Li, W.~Chen, and H.~Li, ``A general differentiable mesh renderer for
  image-based 3d reasoning,'' \emph{IEEE Transactions on Pattern Analysis and
  Machine Intelligence}, 2020.

\bibitem{liu2020neural}
L.~Liu, J.~Gu, K.~Zaw~Lin, T.-S. Chua, and C.~Theobalt, ``Neural sparse voxel
  fields,'' in \emph{Advances in Neural Information Processing Systems
  (Neurips)}, 2020, pp. 15\,651--15\,663.

\bibitem{barron2021mipnerf}
J.~T. Barron, B.~Mildenhall, M.~Tancik, P.~Hedman, R.~Martin-Brualla, and P.~P.
  Srinivasan, ``Mip-nerf: A multiscale representation for anti-aliasing neural
  radiance fields,'' in \emph{Proceedings of the IEEE International Conference
  on Computer Vision (ICCV)}, 2021, pp. 5855--5864.

\bibitem{yu2021pixelnerf}
A.~Yu, V.~Ye, M.~Tancik, and A.~Kanazawa, ``pixelnerf: Neural radiance fields
  from one or few images,'' in \emph{Proceedings of the IEEE Conference on
  Computer Vision and Pattern Recognition (CVPR)}, 2021, pp. 4578--4587.

\bibitem{song2019autoint}
W.~Song, C.~Shi, Z.~Xiao, Z.~Duan, Y.~Xu, M.~Zhang, and J.~Tang, ``Autoint:
  Automatic feature interaction learning via self-attentive neural networks,''
  in \emph{Proceedings of the 28th ACM International Conference on Information
  and Knowledge Management}, 2019, pp. 1161--1170.

\bibitem{grf2020}
A.~Trevithick and B.~Yang, ``Grf: Learning a general radiance field for 3d
  scene representation and rendering,'' in \emph{Proceedings of the IEEE
  International Conference on Computer Vision (ICCV)}, 2021, pp.
  15\,182--15\,192.

\bibitem{henzler2021unsupervised}
P.~Henzler, J.~Reizenstein, P.~Labatut, R.~Shapovalov, T.~Ritschel, A.~Vedaldi,
  and D.~Novotny, ``Unsupervised learning of 3d object categories from videos
  in the wild,'' in \emph{Proceedings of the IEEE Conference on Computer Vision
  and Pattern Recognition (CVPR)}, 2021, pp. 4700--4709.

\bibitem{kendall2015posenet}
A.~Kendall, M.~Grimes, and R.~Cipolla, ``Posenet: A convolutional network for
  real-time 6-dof camera relocalization,'' in \emph{Proceedings of the IEEE
  International Conference on Computer Vision (ICCV)}, 2015, pp. 2938--2946.

\bibitem{wu2017delving}
J.~Wu, L.~Ma, and X.~Hu, ``Delving deeper into convolutional neural networks
  for camera relocalization,'' in \emph{Proceedings of the IEEE International
  Conference On Robotics And Automation (ICRA)}, 2017, pp. 5644--5651.

\bibitem{peng2019pvnet}
S.~Peng, Y.~Liu, Q.~Huang, X.~Zhou, and H.~Bao, ``Pvnet: Pixel-wise voting
  network for 6dof pose estimation,'' in \emph{Proceedings of the IEEE
  Conference on Computer Vision and Pattern Recognition (CVPR)}, 2019, pp.
  4561--4570.

\bibitem{zhou2019continuity}
Y.~Zhou, C.~Barnes, J.~Lu, J.~Yang, and H.~Li, ``On the continuity of rotation
  representations in neural networks,'' in \emph{Proceedings of the IEEE
  Conference on Computer Vision and Pattern Recognition (CVPR)}, 2019, pp.
  5745--5753.

\bibitem{BronsteinBLSV17}
M.~M. Bronstein, J.~Bruna, Y.~LeCun, A.~Szlam, and P.~Vandergheynst,
  ``Geometric deep learning: Going beyond euclidean data,'' \emph{{IEEE} Signal
  Process. Mag.}, vol.~34, no.~4, pp. 18--42, 2017.

\bibitem{KipfW16}
T.~N. Kipf and M.~Welling, ``Semi-supervised classification with graph
  convolutional networks,'' in \emph{International Conference on Learning
  Representations (ICLR)}, 2016.

\bibitem{su15mvcnn}
H.~Su, S.~Maji, E.~Kalogerakis, and E.~G. Learned{-}Miller, ``Multi-view
  convolutional neural networks for 3d shape recognition,'' in
  \emph{Proceedings of the IEEE International Conference on Computer Vision
  (ICCV)}, 2015.

\bibitem{ravi2020pytorch3d}
N.~Ravi, J.~Reizenstein, D.~Novotny, T.~Gordon, W.-Y. Lo, J.~Johnson, and
  G.~Gkioxari, ``Accelerating 3d deep learning with pytorch3d,''
  \emph{arXiv:2007.08501}, 2020.

\bibitem{chang2015shapenet}
A.~X. Chang, T.~Funkhouser, L.~Guibas, P.~Hanrahan, Q.~Huang, Z.~Li,
  S.~Savarese, M.~Savva, S.~Song, H.~Su \emph{et~al.}, ``Shapenet: An
  information-rich 3d model repository,'' \emph{arXiv preprint
  arXiv:1512.03012}, 2015.

\bibitem{insafutdinov2018unsupervised}
E.~Insafutdinov and A.~Dosovitskiy, ``Unsupervised learning of shape and pose
  with differentiable point clouds,'' in \emph{Advances in Neural Information
  Processing Systems (NeurIPS)}, 2018, pp. 2807--2817.

\bibitem{ladicky2017point}
L.~Ladicky, O.~Saurer, S.~Jeong, F.~Maninchedda, and M.~Pollefeys, ``From point
  clouds to mesh using regression,'' in \emph{Proceedings of the IEEE
  International Conference on Computer Vision (ICCV)}, 2017, pp. 3893--3902.

\bibitem{smith19a}
E.~Smith, S.~Fujimoto, A.~Romero, and D.~Meger, ``Geometrics: Exploiting
  geometric structure for graph-encoded objects,'' in \emph{International
  Conference on Machine Learning (ICML)}, 2019, pp. 5866--5876.

\bibitem{curless1996volumetric}
B.~Curless and M.~Levoy, ``A volumetric method for building complex models from
  range images,'' 1996.

\bibitem{LorensenC87}
W.~E. Lorensen and H.~E. Cline, ``Marching cubes: {A} high resolution 3d
  surface construction algorithm,'' in \emph{SIGGRAPH}, 1987.

\bibitem{chen2019dibr}
W.~Chen, H.~Ling, J.~Gao, E.~Smith, J.~Lehtinen, A.~Jacobson, and S.~Fidler,
  ``Learning to predict 3d objects with an interpolation-based differentiable
  renderer,'' \emph{Advances in Neural Information Processing Systems
  (NeurIPS)}, pp. 9609--9619, 2019.

\bibitem{koch2019abc}
S.~Koch, A.~Matveev, Z.~Jiang, F.~Williams, A.~Artemov, E.~Burnaev, M.~Alexa,
  D.~Zorin, and D.~Panozzo, ``Abc: A big cad model dataset for geometric deep
  learning,'' in \emph{Proceedings of the IEEE Conference on Computer Vision
  and Pattern Recognition (CVPR)}, 2019, pp. 9601--9611.

\end{thebibliography}

\begin{IEEEbiography}[{\includegraphics[width=1in,height=1.25in,clip,keepaspectratio]{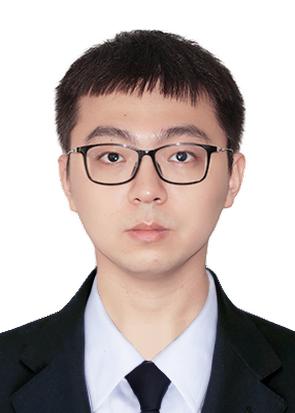}}]{Chao Wen} received the BE degree in computer science from University of Electronic Science and Technology of China, in 2018, the Master degree in computer science from the Academy of Engineering and Technology, Fudan University. His research is focused on 3D shape reconstruction and human image synthesis.
\end{IEEEbiography}

\begin{IEEEbiography}[{\includegraphics[width=1in,height=1.25in,clip,keepaspectratio]{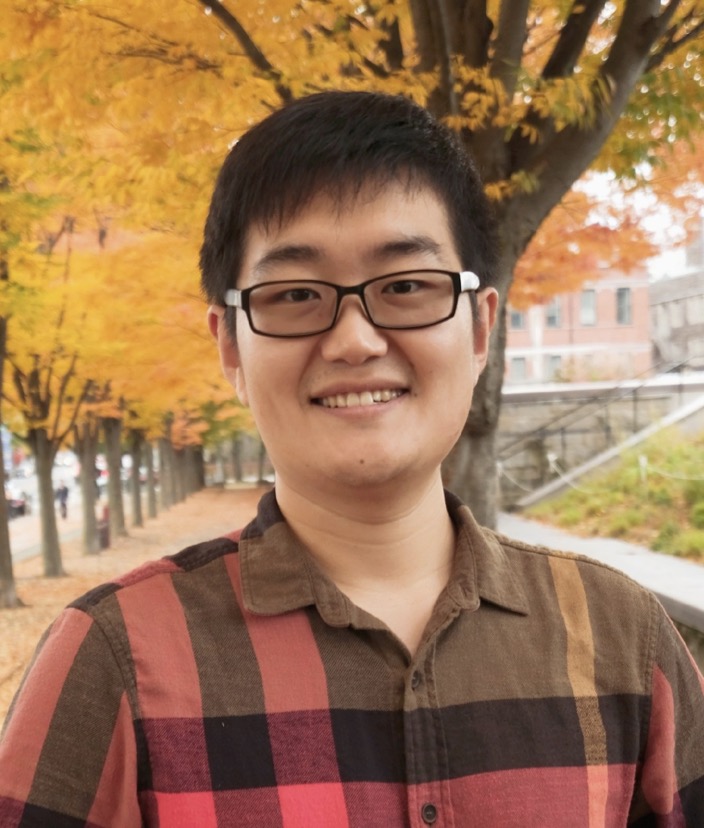}}]{Yinda Zhang} is a Senior Research Scientist at Google. His research interests lie at the intersection of computer vision, computer graphics, and machine learning. Recently, he focuses on empowering 3D vision and perception via machine learning, including dense depth estimation, 3D shape analysis, 3D scene understanding, and neural rendering. He received his Ph.D. in Computer Science from Princeton University, advised by Professor Thomas Funkhouser. Before that, he received a Bachelor degree from Dept. Automation in Tsinghua University, and a Master degree from Dept. ECE in National University of Singapore co-supervised by Prof. Ping Tan and Prof. Shuicheng Yan.
\end{IEEEbiography}

\begin{IEEEbiography}[{\includegraphics[width=1in,height=1.25in,clip,keepaspectratio]{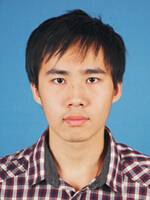}}]{Chenjie Cao} received the M.S. degree in Computer Science from East China University of Science and Technology, in 2019. He is currently pursuing the Ph.D. degree in Statistics from Fudan University. His research interests include machine learning, deep learning, 3D shape reconstruction and computer vision.
\end{IEEEbiography}

\begin{IEEEbiography}[{\includegraphics[width=1in,height=1.25in,clip,keepaspectratio]{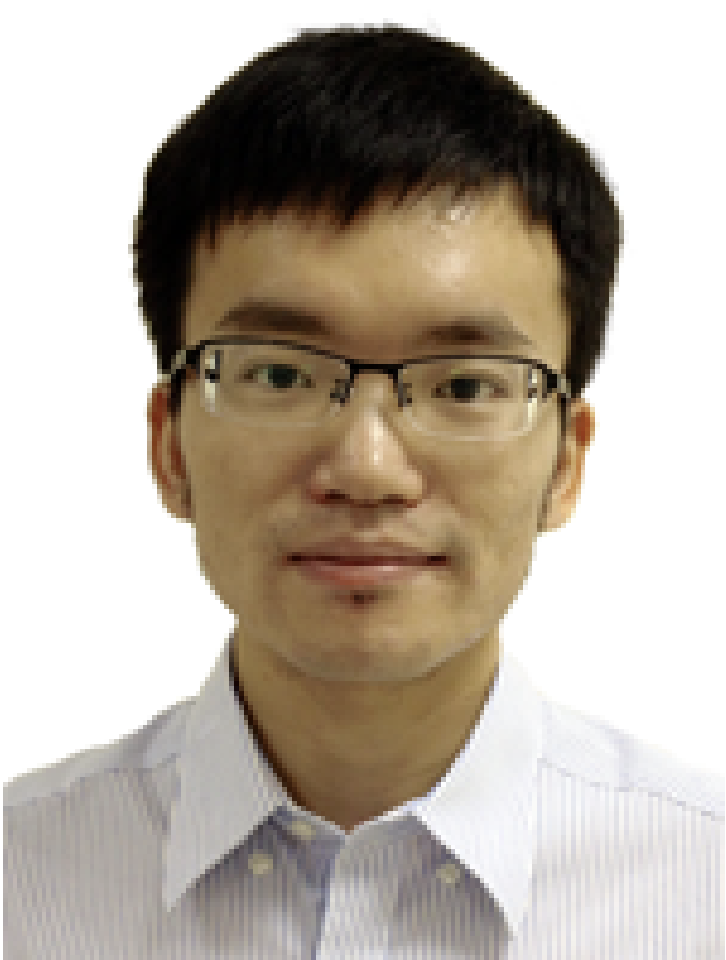}}]{Zhuwen Li} received the BE degree in computer
science from Tianjin University, in 2008, the master’s
degree in computer science from Zhejiang
University, in 2011, and the PhD degree from the
Department of Electrical and Computer Engineering,
National University of Singapore, in 2014.
Currently, he is a research scientist at Nuro, Inc.
Recently, he is working on perception in autonomous
driving, specifically in 3D structure recovery,
motion analysis, point cloud analysis, and
object detection.
\end{IEEEbiography}

\begin{IEEEbiography}[{\includegraphics[width=1in,height=1.25in,clip,keepaspectratio]{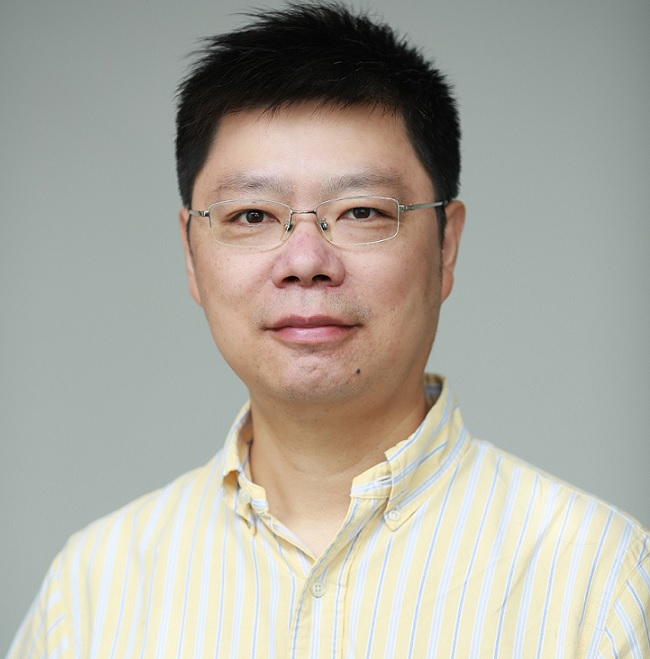}}]{Xiangyang Xue} received the BS, MS, and PhD
degrees in communication engineering from
Xidian University, Xian, China, in 1989, 1992,
and 1995, respectively. He is currently a professor
of computer science with Fudan University,
Shanghai, China. His research interests include
multimedia information processing and machine
learning.
\end{IEEEbiography}

\begin{IEEEbiography}[{\includegraphics[width=1in,height=1.25in,clip,keepaspectratio]{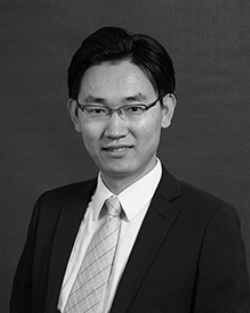}}]{Yanwei Fu} received the MEng degree from the Department of Computer Science and Technology, Nanjing University, China, in 2011, and the PhD degree from the Queen Mary University of London, in 2014. He held a post-doctoral position at Disney Research, Pittsburgh, PA, from 2015 to 2016. He is currently a tenure-track professor with Fudan University.  
He was appointed as the Professor of Special Appointment (Eastern Scholar) at Shanghai Institutions of Higher Learning.
His work has led to many awards, including the IEEE ICME 2019 best paper.
He published more than 100 journal/conference papers including IEEE TPAMI, TMM, ECCV, and CVPR. His research interests are one-shot learning, and learning based 3D reconstruction.
\end{IEEEbiography}

\end{document}